%% file: rewardverse_arxiv.tex
\documentclass{article} % For LaTeX2e
\usepackage{iclr2027_conference,times}

\input{math_commands.tex}

\usepackage{hyperref}
\usepackage{url}
\usepackage{booktabs}
\usepackage{graphicx}
\usepackage{amsmath}
\usepackage{amssymb}
\usepackage{bm}
\usepackage{multirow}
\usepackage{pifont}
\usepackage{algorithm}
\usepackage{algorithmic}

\usepackage[most]{tcolorbox}
\usepackage{xcolor}

\newcommand{\xmark}{\ding{55}}

\definecolor{promptbg}{RGB}{247,247,247}
\definecolor{promptframe}{RGB}{205,208,211}

\newtcblisting{promptbox}{
  enhanced,
  breakable,
  listing only,
  colback=promptbg,
  colframe=promptframe,
  boxrule=0.5pt,
  arc=1.5pt,
  left=6pt,
  right=6pt,
  top=5pt,
  bottom=5pt,
  before skip=6pt,
  after skip=8pt,
  listing options={
    basicstyle=\ttfamily\footnotesize,
    breaklines=true,
    columns=fullflexible,
    keepspaces=true,
    showstringspaces=false
  }
}

\title{RewardVerse: Rubric-Guided Policy Optimization for Video Reward Modeling}

\author{%
{\bfseries
Zhenchen Tang$^{1,2,4}$ \quad
Yang Li$^{1,2,4}$ \quad
Songlin Yang$^{3,4,\dagger}$ \quad
Bo Peng$^{1,2}$} \\[4pt]
{\bfseries
Xiaotong Zhao$^{4}$ \quad
Shuai Li$^{4}$ \quad
Haotian Fan$^{4}$ \quad
Alan Zhao$^{4}$ \quad
Jing Dong$^{1,2,*}$} \\[7pt]
{\normalfont\small
$^{1}$New Laboratory of Pattern Recognition, Institute of Automation, Chinese Academy of Sciences} \\
{\normalfont\small
$^{2}$School of Artificial Intelligence, University of Chinese Academy of Sciences} \\
{\normalfont\small
$^{3}$The Hong Kong University of Science and Technology} \\
{\normalfont\small
$^{4}$Tencent} \\[3pt]
{\normalfont\small
$^{\dagger}$Project Lead \qquad
$^{*}$Corresponding Author}
}

\iclrfinalcopy % Non-anonymous (arXiv preprint) layout: shows the author block.

\hypersetup{
  colorlinks=true,
  linkcolor=blue!55!black,
  citecolor=blue!55!black,
  urlcolor=blue!65!black
}

\begin{document}

\maketitle

% Preprint running header: the ICLR-style header is removed for the arXiv version.
\lhead{}
\rhead{}

\begin{abstract}
Reinforcement learning (RL) is vital for optimizing video generation models, with a robust reward model (RM) serving as the cornerstone. However, existing video reward models often produce unstable scalar scores because they directly map complex, subjective video quality into a single score without explicit evaluation criteria. This leads to \emph{scalar drift}, where the scoring scale collapses or shifts across different prompts, making the reward unreliable for RL. Drawing inspiration from professional human annotation engineering, we address this problem with \textbf{RewardVerse}, a rubric-based video reward framework that introduces a \emph{dynamic rubric} as an intermediate representation between the evaluation query and the scorer. Instead of unconstrained direct scoring, RewardVerse first generates explicit evaluation criteria and then performs rubric-guided scoring, providing a stable semantic anchor that mitigates scalar drift. To efficiently optimize this collaborative pipeline, we propose \textbf{R}ubric-\textbf{G}uided \textbf{P}olicy \textbf{O}ptimization (\textbf{RGPO}), a two-stage training algorithm. RGPO first warms up the scorer using self-evolving seed rubrics and then jointly optimizes the rubric generator to produce query-adaptive evaluation criteria while continuously aligning the scorer with human ratings. Extensive experiments on the 16-dimensional EvalVerse benchmark and external datasets demonstrate that RewardVerse mitigates scalar drift, achieves state-of-the-art performance on both pointwise and pairwise evaluation, and provides a robust and interpretable reward signal for RL in video generation.
\end{abstract}

\section{Introduction}
\label{sec:intro}
Generative video models have achieved remarkable progress in recent years. As these models continue to improve, reinforcement learning (RL) has become an important paradigm for aligning generated videos with human preferences, with a robust reward model (RM) serving as the cornerstone. 
Despite recent progress, existing video RMs still struggle to produce stable pointwise rewards. Most methods directly map a generated video to a single scalar score, either through discriminative regression or generative reasoning. Since video quality is inherently subjective and multi-dimensional~\citep{wang2026cinetechbench}, this direct scoring process lacks explicit evaluation criteria. Consequently, the scoring scale often becomes unstable, causing scores to collapse into a narrow range \citep{jin2026z} or shift under different prompts and contexts. We refer to this phenomenon as \emph{scalar drift}.

To address this challenge, we draw inspiration from professional human annotators, who rarely assign scores directly. Instead, they first decompose an evaluation task into explicit criteria before producing a final judgment, creating a stable semantic anchor that maintains consistency across different samples \citep{tong2025mj, xu2026visionreward}. Existing video reward models largely omit this explicit criterion-setting stage. By forcing a model to map queries to scores in an unconstrained step, their internal scoring standards shift across different prompts, leading to unstable rewards.

\begin{figure}[t]
\centering
\includegraphics[width=1.0\linewidth]{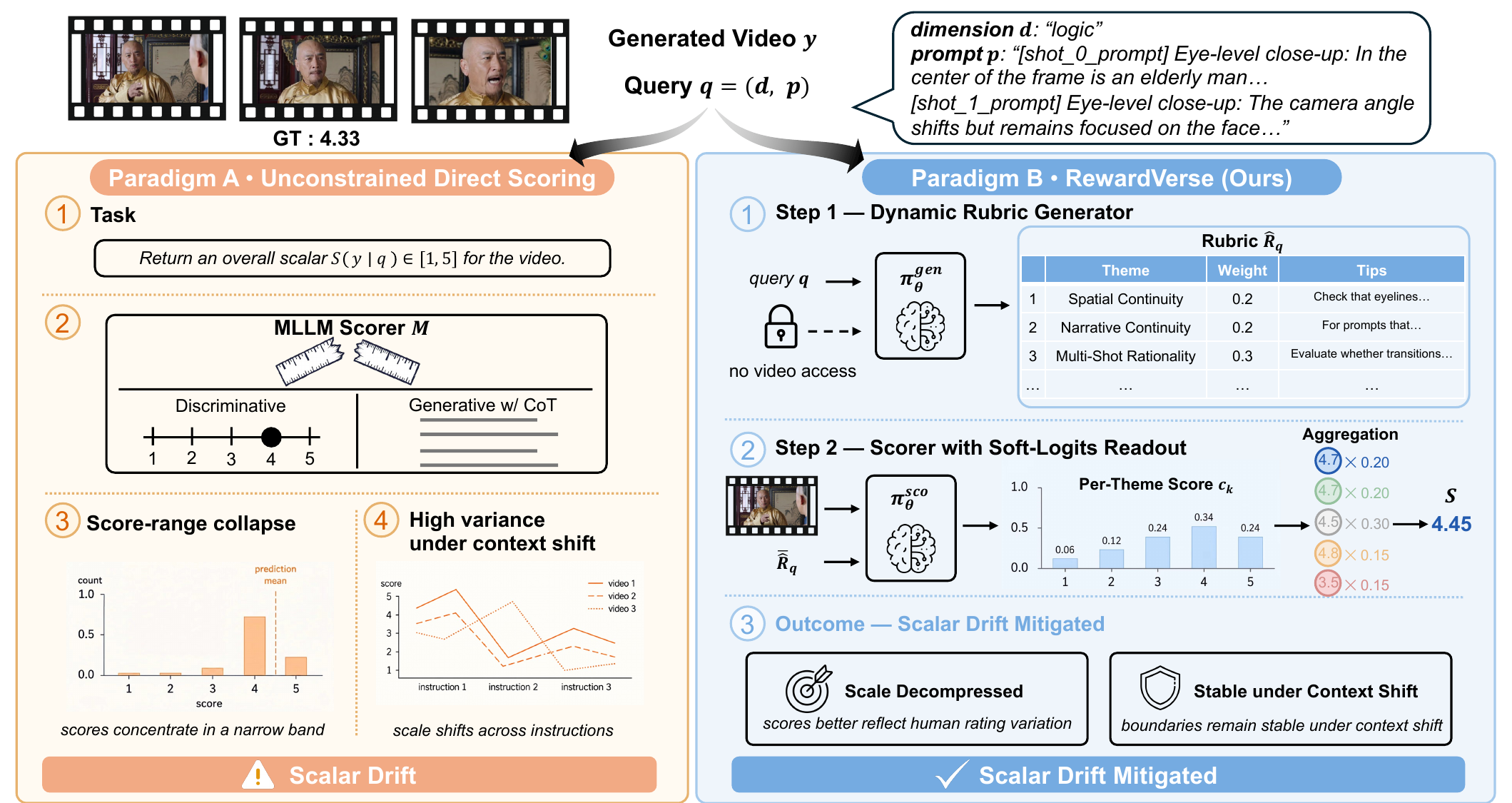}
\vspace{-0.6cm}
\caption{A comparison of different video evaluation paradigms. Compared to unconstrained discriminative and generative scoring that suffer from severe scalar drift, RewardVerse inserts a dynamic rubric as an intermediate representation to anchor the scoring process.}
\label{fig:paradigms}
\vspace{-0.5cm}
\end{figure}

Based on this insight, we propose \textbf{RewardVerse}, a highly data-efficient video evaluation framework that natively supports both pointwise and pairwise evaluation. As compared in Figure~\ref{fig:paradigms}, instead of directly mapping a video to a score, we insert a \emph{dynamic rubric} as an intermediate representation to decouple evaluation into standard generation and objective execution. Specifically, the rubric generator first decomposes the holistic text query into explicit evaluation themes, weights, and scoring tips. The scorer then evaluates the video against these rubrics independently, rather than relying on an implicit internal standard. This decoupling provides a clear cognitive anchor that mitigates score-range collapse and improves robustness under context shift.

To efficiently learn dynamic rubrics and optimize the collaborative pipeline with minimal preference pairs and without prior supervised fine-tuning (SFT), we introduce a two-stage training paradigm called \textbf{Rubric-Guided Policy Optimization} (\textbf{RGPO}) under Group Relative Policy Optimization (GRPO). We first warm up the scorer using self-evolving seed rubrics synthesized offline. Then, we jointly optimize the rubric generator to produce query-adaptive evaluation criteria while continuously aligning the scorer with human ratings. Through this optimization, RewardVerse effectively mitigates scalar drift, requiring only 30 preference pairs per dimension to deliver state-of-the-art performance and provide a robust, highly interpretable reward for downstream generation models.

Our main contributions are summarized as follows:
\begin{itemize}
    \item \textbf{Rubric-as-Reward for Video RMs:} To our knowledge, we are the first to introduce the rubric-as-reward paradigm to video reward modeling. We propose \textbf{RewardVerse}, a framework that employs a dynamic rubric as an intermediate representation to stabilize pointwise scores and mitigate scalar drift. 
    \item \textbf{Data-Efficient Joint Policy Optimization:} We design \textbf{RGPO} (Rubric-Guided Policy Optimization), a two-stage training algorithm that learns a dynamic rubric generator from only 30 preference pairs per dimension, while maintaining a human-aligned scorer.
    \item \textbf{Systematic Empirical Validation:} We conduct extensive evaluations on the multi-dimensional EvalVerse-adapted dataset and external benchmarks. RewardVerse achieves state-of-the-art evaluation performance and serves as a stable, interpretable reward for downstream video generation optimization.
\end{itemize}

\section{Related Work}
\label{sec:related}

\paragraph{The Rubric-as-Reward Paradigm.}
Rubrics decompose quality evaluation into explicit criteria corresponding to concrete aspects of the desired output, providing a natural handle for open-ended, non-verifiable tasks. 
In LLM and autonomous agent domains, recent works bring rubrics into RL for subjective tasks such as medicine, writing, and research (e.g., RaR~\citep{gunjal2025rar}, RLCF~\citep{viswanathan2026checklists}, HealthBench~\citep{arora2025healthbench}, RLCER~\citep{sheng2026reinforcing}, OpenRubrics~\citep{liu2026openrubrics}, Rubric-ARM~\citep{xu2026alternating}, and DR Tulu~\citep{shao2025dr}). 
Similarly, in image generation and editing, recent works use explicit rubrics, checklists, or decomposed criteria to evaluate and optimize visual outputs (e.g., AlphaGRPO~\citep{huang2026alphagrpo}, Edit-R1~\citep{guo2026leveraging}, EditReward-Compass~\citep{bai2026edit}, ARR-RPO~\citep{tian2026auto}, and RewardHarness~\citep{zhang2026rewardharness}). 
These works establish explicit criteria as useful reward signals across open-ended generation tasks. 
In contrast, RewardVerse makes the rubric a dynamic, learned intermediate representation: its themes, weights, and scoring tips are generated from each evaluation query without access to the candidate video, and are jointly optimized with the scorer to stabilize pointwise video rewards.

\paragraph{Video Reward Modeling.}
Existing video reward models generally fall into two categories: discriminative and generative. 
Discriminative models (e.g., VideoScore~\citep{he2024videoscore}, VideoReward~\citep{liu2026improving}) train regression layers on top of visual encoders to output continuous scalar scores. 
Generative models (e.g., UnifiedReward~\citep{unifiedreward}, VideoScore2~\citep{he2025videoscore2}) prompt multi-modal LLMs to output scores directly. 
Because these models score videos directly without explicit anchors, they often suffer from scalar drift. 
As analyzed in Section~\ref{sec:phenomenon}, scalar drift causes scores to collapse into a narrow range or change wildly across prompts, making them unstable for online RL training. By jointly training rubric generation and scoring under our structured rubric bottleneck, RewardVerse mitigates scalar drift, achieves high data efficiency, and provides a stable reward signal for downstream video generation models.

\section{The Phenomenon of Scalar Drift in Video Evaluation}
\label{sec:phenomenon}

To establish the empirical foundations of our work, we conduct an analytical study of video evaluation behaviors in modern Multi-modal Large Language Models (MLLMs). We focus on dissecting scalar drift under unconstrained direct scoring, analyzing how rubric-guided scoring alleviates this problem, and identifying the remaining limitations that motivate joint policy optimization.

\subsection{Defining Scalar Drift}
\label{sec:scalar-drift-def}
We consider an MLLM $\mathcal{M}$ tasked with returning a single overall scalar reward $S(y\mid q)\in[1,5]$ for a generated video $y$ given a textual query $q$. Because video quality assessment is multi-dimensional and largely subjective, this direct task is \emph{unconstrained}: the model has no explicit anchor for what each point on the 1--5 scale should mean. Under this setting, the internal scoring rubric drifts in two interacting ways, which together we define as \textbf{scalar drift}:
\begin{itemize}
    \item \textbf{Score-range collapse.} Pointwise scores concentrate in a narrow ``safe'' high band, shrinking the score gap between preferred and non-preferred videos and leaving downstream RL with near-flat gradients.
    \item \textbf{High variance under context shift.} Without explicit anchors, the scoring scale may shift across paraphrased instructions or different input orderings, producing inconsistent scores or pairwise decisions for the same content.
\end{itemize}
A workable pointwise reward must therefore (a) preserve scale resolution across the population of videos, and (b) be robust to context phrasing and ordering. See Appendix~\ref{app:scalar_drift} for further analysis and Appendix~\ref{app:cross_rm_drift} for score distributions across existing reward models.

\subsection{Empirical Study: Rubric is Necessary but Not Sufficient}
\label{sec:variants}
To isolate which design choice cures which symptom of scalar drift, we construct a $2\times 2$ ablation design space covering rubric injection (presence vs. absence) and decoding format (natural-language generation vs. soft-logits continuous expectation). This yields four variants evaluated on a hold-out pool of $240$ pointwise videos with human labels (details in Appendix~\ref{app:full-ablation}): 1) \textbf{V1 (No Rubric + Natural Float)}; 2) \textbf{V2 (No Rubric + Soft-Logits)}; 3) \textbf{V3 (With Rubric + Soft-Logits, Deployed)}; and 4) \textbf{V4 (With Rubric + Natural Float)}. Figure~\ref{fig:scalar_drift_hist} reveals three key insights that drive our downstream framework design:

\begin{figure}[t]
\centering
\includegraphics[width=0.95\linewidth]{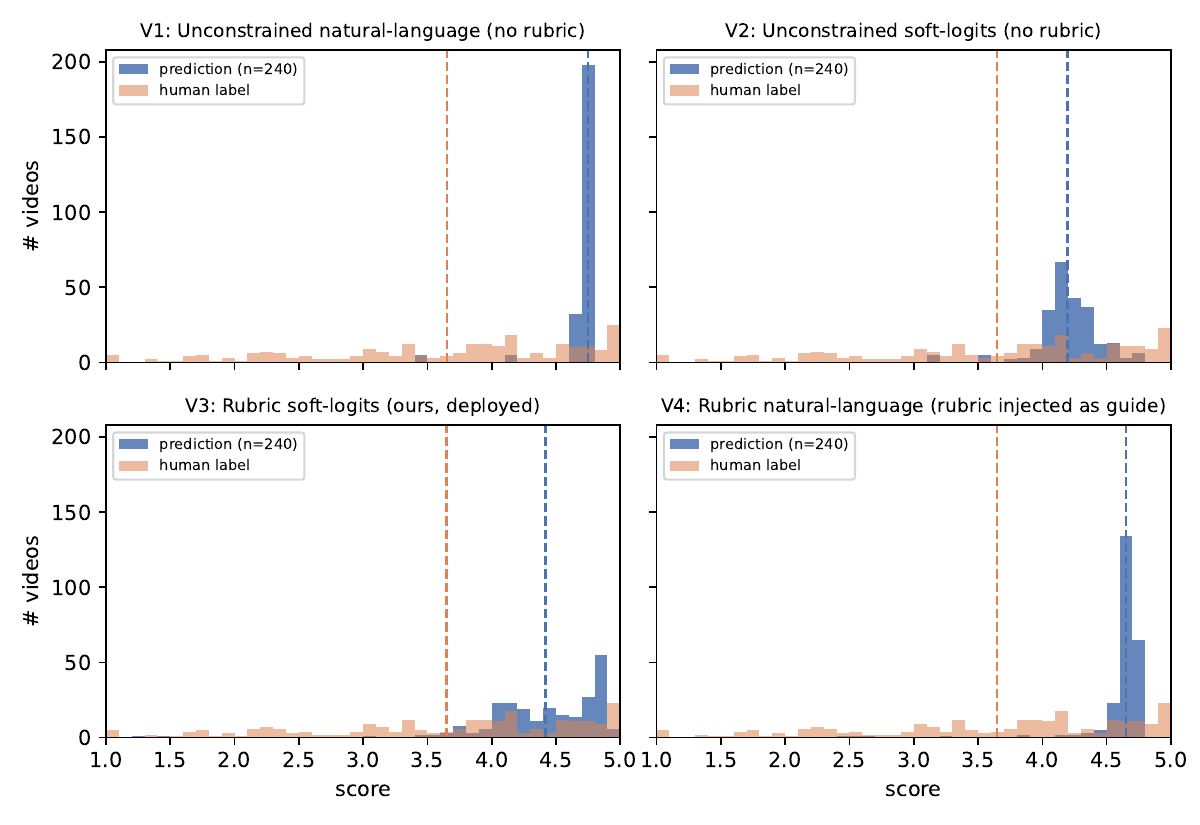}
\vspace{-0.6cm}
\caption{Pointwise score distributions of four scoring variants. Blue bars are predictions, orange bars are human labels, and dashed lines show the means. \textbf{V1} and \textbf{V2} show score-range collapse, while \textbf{V4} remains affected by natural-language bias. Only \textbf{V3} (rubrics with soft-logits) effectively spreads predictions, but it still deviates from human ratings, motivating RGPO in Section~\ref{sec:method}.}
\label{fig:scalar_drift_hist}
\vspace{-0.5cm}
\end{figure}

\textbf{Cognitive Anchoring via Rubrics.} Without rubric guidance, unconstrained direct scoring (V1 and V2) suffers from severe score saturation and narrow variance. Injecting a rubric as a cognitive anchor substantially expands these distributions. For example, holding the soft-logits format fixed, adding the rubric ($V2 \to V3$) expands the scoring standard deviation by \textbf{$+82.6\%$} ($\sigma$: $0.258 \to 0.471$) for Qwen2.5-VL-7B. Under natural text generation, the rubric acts as a vital regularizer: it expands the collapsed standard deviation of Qwen2.5-VL-7B ($V1 \to V4$: $0.203 \to 0.275$), while reducing the excessively noisy and unguided variance of Gemini-3.1-Pro (reducing $\sigma$ from $1.500$ to $1.158$), bringing their score variance closer to that of human judgments (see Appendix~\ref{app:gemini-1k}).

\textbf{Continuous Scale Decompression via Soft-Logits.} Crucially, however, the rubric \emph{alone} is insufficient if paired with natural text generation. In V4, two-thirds of the comparison pairs still tie because natural-language outputs remain anchored to the model's high-band linguistic prior. Further mitigating scalar drift requires coupling the rubric with a continuous, logit-based readout mechanism. Extracting a continuous expected score directly from the logits of rating tokens (soft-logits) successfully bypasses the model's generation bias (formulated in Section~\ref{sec:method}). Consequently, pairing a rubric with soft-logits (V3) better captures the variation in human ratings and reduces score-range collapse. For our deployment, we choose this pointwise multi-theme soft-logits protocol (V3) as our primary configuration; detailed protocol comparisons are provided in Appendix~\ref{app:protocols}.

\textbf{The Remaining Limitations.} Although rubric-guided soft-logits scoring mitigates scalar drift, two limitations remain. First, static rubrics cannot adapt their themes, tips, and weights to different queries. Second, the predicted score distribution may still deviate
from human ratings, even for stronger proprietary MLLMs such as Gemini-3.1-Pro (Appendix~\ref{app:gemini-1k}). These results show that rubric prompting alone cannot fully resolve scalar drift, motivating RGPO to jointly optimize rubric generation and scoring. As verified in Appendix~\ref{app:post_rgpo_drift}, the complete RGPO training further improves both score alignment and correlation with human ratings.

\section{Methodology}
\label{sec:method}

Rather than treating the rubric as a fixed prompt scaffold, we make it a learnable intermediate representation. This allows the model to generate query-adaptive evaluation criteria while keeping the scorer aligned with human ratings. This joint optimization further mitigates the scalar drift that remains after rubric-guided scoring.
Building on this formulation, we present \textbf{RewardVerse}, which parameterizes adaptive rubric generation and human-aligned scoring within a dual-role policy model $\pi_\theta$. We optimize the model via \textbf{Rubric-Guided Policy Optimization (RGPO)}, a two-stage training algorithm. Stage 1 warms up the scorer using self-evolving seed rubrics (§\ref{sec:warmup}), and Stage 2 optimizes the dynamic rubric generator while continuing to calibrate the scorer (§\ref{sec:joint}).

\begin{figure}[t]
\centering
\includegraphics[width=1.0\linewidth]{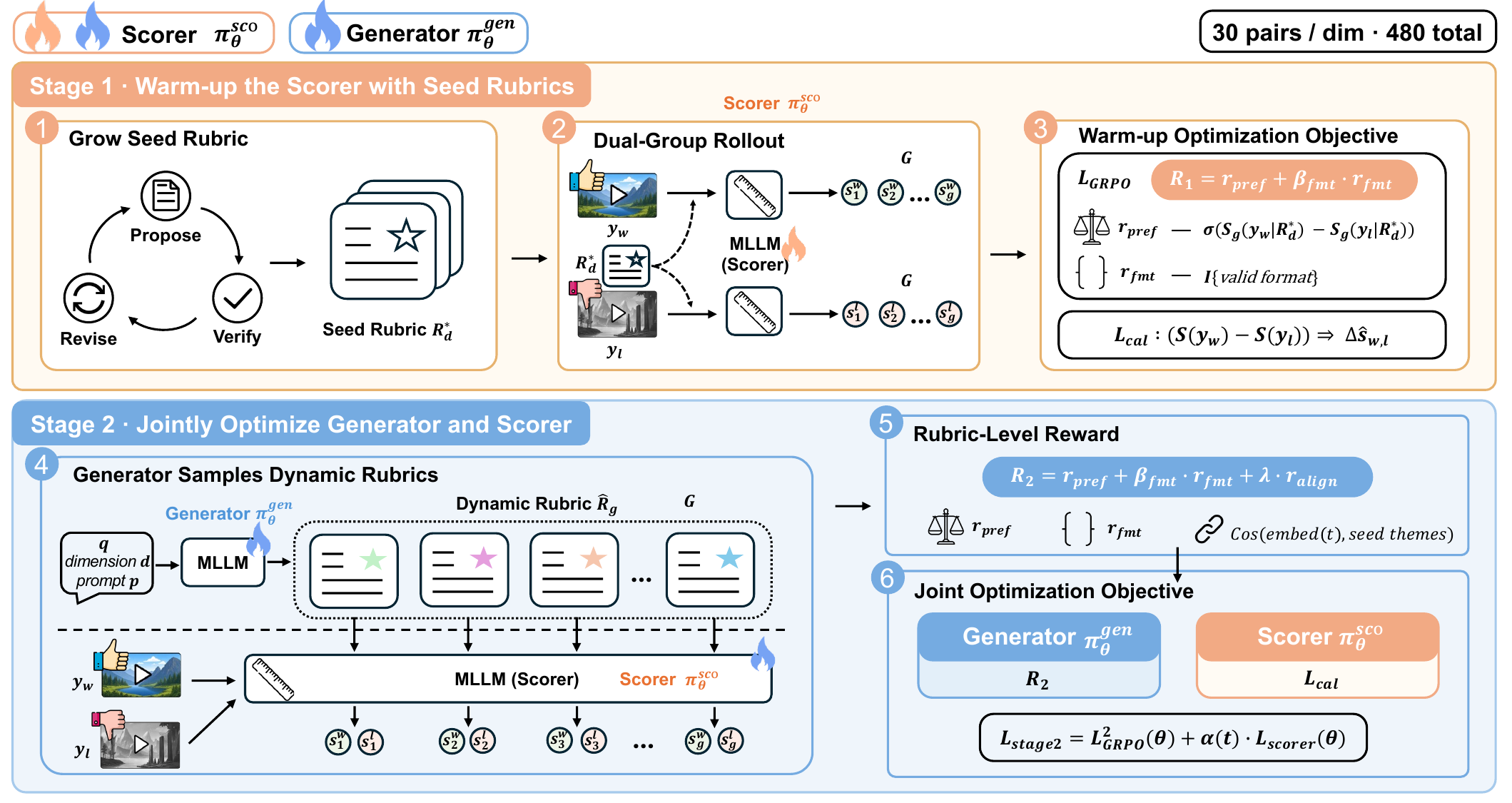}
\vspace{-0.6cm}
\caption{Overview of RewardVerse and its two-stage RGPO training. Stage 1 warms up the scorer with seed rubrics using preference and format rewards plus a calibration loss. Stage 2 learns the dynamic rubric generator while continuing scorer calibration.}
\label{fig2}
\vspace{-0.5cm}
\end{figure}

\subsection{RewardVerse Pipeline and Formulation}
\label{sec:pipeline}

\paragraph{Inputs.}
An evaluation query $q = (d, p)$ pairs a target dimension $d \in \mathcal{D}$ with the text prompt $p$ used to synthesize the candidate video $y$. Each video is scored independently during both training and inference, avoiding the pairwise position bias analyzed in Appendix~\ref{app:perturbation}.

\paragraph{Module 1 --- Dynamic Rubric Generator $\pi_\theta^{\mathrm{gen}}$.}
The policy first acts as a generator that parses the query $q$ and emits a dynamic rubric under a fixed schema:
  \begin{equation}
      \hat{R}_q \sim \pi_\theta^{\mathrm{gen}}(\cdot \mid q),
      \qquad
      \hat{R}_q = \bigl\{(t_k, w_k, T_k)\bigr\}_{k=1}^{K},
      \label{eq:rubric_gen}
  \end{equation}
where $t_k$ is a theme, $w_k \in (0,1)$ is its weight with $\sum_k w_k = 1$, and $T_k$ is a set of execution tips based on queries. 

\textit{Candidate-Independent Rubric Generation:} Notably, the generator takes only the query $q$ as input without seeing the video $y$. This decoupling prevents the generator from producing biased, video-dependent rubrics (e.g., being overly lenient to low-quality videos or overly strict to high-quality ones), safeguarding evaluation objectivity while keeping the criteria query-adaptive.

\paragraph{Module 2 --- Scorer $\pi_\theta^{\mathrm{sco}}$ with Soft-Logits Readout.}
Given a rubric $R$, the policy emits one score per theme. To avoid parsing digits under JSON-constrained decoding, at each score slot we read the logits $l_v$ of the five rating tokens $v \in \{1,\dots,5\}$ (corresponding to the string representations of integers ``1'' to ``5'' in the vocabulary) and take the expected value:
  \begin{equation}
      c_k(y \mid R)
      \;=\; \sum_{v=1}^{5} v \cdot P_\theta\!\bigl(v \mid y, t_k, T_k\bigr),
      \qquad
      P_\theta\!\bigl(v \mid y, t_k, T_k\bigr)
      \;=\; \frac{\exp(l_v)}{\sum_{v'=1}^{5}\exp(l_{v'})}.
      \label{eq:soft_logits}
  \end{equation}
The pointwise reward aggregates theme scores by their weights:
  \begin{equation}
      S(y \mid R) \;=\; \sum_{k=1}^{K} w_k \cdot c_k(y \mid R).
      \label{eq:score_agg}
  \end{equation}
The generator and scorer share the same backbone and are distinguished by role-specific prompts. In Stage 2, the shared policy is trained with separate signals for the two roles. The main prompt templates are provided in Appendix~\ref{app:prompts}.

\subsection{Stage 1: Seed-Guided Scorer Warm-up}
\label{sec:warmup}

Learning dynamic rubric generation and calibrated scoring at the same time from scratch is highly under-determined. A poor generator produces chaotic rubrics that disrupt scorer learning. Therefore, Stage 1 uses fixed seed rubrics $R^\star_d$ and updates the shared policy through the scorer role, establishing human-aligned score margins before dynamic rubric generation is introduced.

\paragraph{Seed Rubric Self-Evolving.}
For each dimension $d$, we build a seed rubric $R^\star_d$ through an offline loop driven by a frontier MLLM $\mathcal{M}$ (Gemini-3.1-Pro) over $30$ preference pairs per dimension:
  \begin{itemize}
      \item \textbf{Propose ($\mathcal{M}^{\mathrm{gen}}$).} Given $(q, y_w, y_l)$, $\mathcal{M}$ drafts a candidate rubric $\tilde{R}$ that explains why $y_w \succ y_l$.
      \item \textbf{Verify ($\mathcal{M}^{\mathrm{ver}}$).} $\mathcal{M}$ scores both videos with $\tilde{R}$; $\tilde{R}$ is accepted only if it recovers $y_w \succ y_l$.
      \item \textbf{Revise ($\mathcal{M}^{\mathrm{ref}}$).} On failure, $\mathcal{M}$ refines $\tilde{R}$ using a critique of the misalignment.
  \end{itemize}
Verified rubrics are pooled into $\mathcal{V}_d$, deduplicated by a two-level Jaccard filter, and reduced to $M{=}5$ representative entries by an $\mathrm{MCR}^2$ sampler~\citep{yu2020learning}, giving $R^\star_d = \{r^\star_{d,1}, \dots, r^\star_{d,M}\}$.

\paragraph{Scorer Optimization via Policy Gradient.}
For each pair $(q, y_w, y_l) \sim \mathcal{D}_{\mathrm{train}}$ under the seed rubric $R^\star_d$, the scorer produces $G$ joint scoring samples. Each sample contains one completion $a_{w,g}$ for the preferred video and one completion $a_{l,g}$ for the non-preferred video, sampled from $\pi_{\theta_{\mathrm{old}}}^{\mathrm{sco}}$ for $g=1,\dots,G$.
Each completion pair $g$ is scored by a pairwise preference-driven reward:
\begin{equation}
    r_{\mathrm{pref}}(a_{w,g}, a_{l,g})
    = \sigma\!\bigl(S_g(y_w \mid R^\star_d) - S_g(y_l \mid R^\star_d)\bigr),
    \label{eq:stage1_pref}
\end{equation}
where $\sigma(\cdot)$ is the sigmoid function, and $S_g$ represents the pointwise score (\eqref{eq:score_agg}) decoded from the corresponding completion $a_{x,g}$ ($x \in \{w,l\}$). The full warm-up reward is defined as:
\begin{equation}
    \mathcal{R}_1(a_{w,g}, a_{l,g})
    = r_{\mathrm{pref}}(a_{w,g}, a_{l,g})
    + \beta_{\mathrm{fmt}}\left(r_{\mathrm{fmt}}(a_{w,g}) + r_{\mathrm{fmt}}(a_{l,g})\right),
    \label{eq:stage1_reward}
\end{equation}
where $\beta_{\mathrm{fmt}}$ controls the contribution of the format reward, and $r_{\mathrm{fmt}}(a_{x,g}) \in \{0,1\}$ is a binary indicator checking whether the completion conforms to the requested scoring format.

Following GRPO~\citep{shao2024deepseekmath}, the joint reward of each pair is standardized across the group to yield the trajectory-level advantage $\hat{A}_g = \bigl(\mathcal{R}_1(a_{w,g}, a_{l,g}) - \mathrm{mean}(\mathcal{R}_1)\bigr)/\mathrm{std}(\mathcal{R}_1)$, which is broadcast to all tokens in both $a_{w,g}$ and $a_{l,g}$. Denoting the per-token importance ratio between the current and behavior scorers by $\rho_{x,g,t}(\theta) = \frac{\pi_\theta^{\mathrm{sco}}(a_{x,g,t}\mid y_x, R^\star_d)}{\pi_{\theta_{\mathrm{old}}}^{\mathrm{sco}}(a_{x,g,t}\mid y_x, R^\star_d)}$, the Stage-1 GRPO objective optimizes both preferred and non-preferred completions:
\begin{equation}
\begin{split}
    \mathcal{L}_{\mathrm{GRPO}}^{(1)}(\theta)
    = -\frac{1}{G}\sum_{g=1}^{G}\frac{1}{2}\sum_{x \in \{w,l\}}\frac{1}{|a_{x,g}|}\sum_{t=1}^{|a_{x,g}|}
    \Bigl[ &\min\!\bigl(\rho_{x,g,t}(\theta)\hat{A}_g,\,
    \mathrm{clip}(\rho_{x,g,t}(\theta),1{-}\epsilon,1{+}\epsilon)\hat{A}_g\bigr) \\
    &-\beta_{\mathrm{KL}}\,\mathbb{D}_{\mathrm{KL}}\!\bigl[\pi_\theta^{\mathrm{sco}} \,\big\|\, \pi_{\mathrm{ref}}^{\mathrm{sco}}\bigr]_{x,g,t} \Bigr].
\end{split}
\label{eq:stage1_grpo}
\end{equation}
Here, $\epsilon$ is the clipping coefficient, $\beta_{\mathrm{KL}}$ controls the KL penalty, and $D_{\mathrm{KL}}[\cdot \| \cdot]_{x,g,t}$ is the per-token KL penalty against a frozen reference policy $\pi^{\mathrm{sco}}_{\mathrm{ref}}$.

\paragraph{Human-Aligned Margin Calibration.}
Although the preference reward encourages the scorer to rank the preferred video above the non-preferred one, it does not control the magnitude of the predicted score difference. We therefore add a calibration loss based on the human score margin:
\begin{equation}
    \mathcal{L}_{\mathrm{cal}}(\theta \mid R)
    = \frac{1}{S_{\max}-S_{\min}}
    \Bigl|\bigl(S(y_w \mid R) - S(y_l \mid R)\bigr) - \Delta\hat{s}_{w,l}\Bigr|,
    \label{eq:cal_loss}
\end{equation}
where $S_{\max}=5$, $S_{\min}=1$, and $\Delta\hat{s}_{w,l} \in [0,4]$ is the target score margin derived from human scores. This loss calibrates the predicted score difference against the human-annotated margin, preventing arbitrary compression or inflation of reward gaps. We focus on relative rather than absolute calibration, since reward models mainly require reliable preference ordering and meaningful score differences, while direct absolute score regression can easily overfit in our low-data setting. The overall Stage-1 objective combines the GRPO loss with this calibration term:
\begin{equation}
    \mathcal{L}_{\mathrm{stage1}}(\theta)
    = \mathcal{L}_{\mathrm{GRPO}}^{(1)}(\theta)
    + \lambda_{\mathrm{cal}}\,\mathcal{L}_{\mathrm{cal}}(\theta \mid R^\star_d).
    \label{eq:stage1_loss}
\end{equation}

\subsection{Stage 2: Joint Policy Optimization via RGPO}
\label{sec:joint}

Stage 2 learns the rubric generator to produce evaluation criteria tailored to each query. For each triple $(q, y_w, y_l)$, the generator samples $G$ dynamic rubrics $\{\hat{R}_g\}_{g=1}^{G} \sim \pi_{\theta_{\mathrm{old}}}^{\mathrm{gen}}(\cdot \mid q)$. Each rubric is judged by how well its induced pointwise scores separate the preferred and non-preferred videos:
\begin{equation}
    r_{\mathrm{pref}}(\hat{R}_g)
    = \sigma\!\bigl(S_g(y_w \mid \hat{R}_g) - S_g(y_l \mid \hat{R}_g)\bigr).
    \label{eq:stage2_pref}
\end{equation}

To keep the generated rubric aligned with the target dimension and enforce the required output format, we define the Stage-2 reward as:
\begin{equation}
    \mathcal{R}_2(\hat{R}_g)
    = r_{\mathrm{pref}}(\hat{R}_g)
    + \beta\,r_{\mathrm{fmt}}(\hat{R}_g)
    + \lambda\,r_{\mathrm{align}}(\hat{R}_g),
    \label{eq:stage2_reward}
\end{equation}
where $r_{\mathrm{align}}(\hat{R}_g)$ measures the cosine similarity between the BGE-M3 embeddings~\citep{chen2024bge} of the generated and seed themes. Here, $\lambda$ and $\beta$ are fixed coefficients for rubric alignment and format regularization, respectively. $r_{\mathrm{fmt}}(\hat{R}_g)$ checks whether the output conforms to the structured JSON schema and whether the rubric weights sum to one.

The composite reward of each generated rubric is standardized across the group to yield the rubric-level advantage $\hat{A}_g = \bigl(\mathcal{R}_2(\hat{R}_g) - \mathrm{mean}(\mathcal{R}_2)\bigr)/\mathrm{std}(\mathcal{R}_2)$, which is broadcast to all tokens in the generated rubric $\hat{R}_g$. The generator is optimized using the same clipped GRPO objective as in \eqref{eq:stage1_grpo}, with $\hat{R}_g$ as the optimized sequence and $\rho_{g,t}(\theta) = \frac{\pi_\theta^{\mathrm{gen}}(\hat{R}_{g,t}\mid q)}{\pi_{\theta_{\mathrm{old}}}^{\mathrm{gen}}(\hat{R}_{g,t}\mid q)}$ as the per-token importance ratio. We denote this generator objective by $\mathcal{L}_{\mathrm{GRPO}}^{(2)}(\theta)$; its full form is provided in Appendix~\ref{loss}. As the rubric distribution changes during Stage 2, we retain scorer calibration to keep the scores aligned with human margins under the sampled rubrics.

\paragraph{Asymmetric Optimization Signals.}
If the scorer were optimized directly by the relative reward $\mathcal{R}_2$, it could increase the reward simply by enlarging the score difference between the preferred and non-preferred videos, without improving score calibration. We therefore use different optimization signals for the two roles. The generator is optimized by the rubric-level GRPO objective, while the scorer receives no direct policy-gradient update from $\mathcal{R}_2$ and is instead optimized by the margin calibration loss. The generated rubric is treated as fixed text context when computing the scorer loss, so the calibration loss does not backpropagate through the rubric:
\begin{equation}
    \mathcal{L}_{\mathrm{scorer}}(\theta)
    = \frac{1}{G}\sum_{g=1}^{G}
    \mathcal{L}_{\mathrm{cal}}\!\bigl(\theta \mid \hat{R}_g\bigr).
    \label{eq:stage2_scorer}
\end{equation}
The overall Stage-2 objective is
\begin{equation}
    \mathcal{L}_{\mathrm{stage2}}(\theta)
    = \mathcal{L}_{\mathrm{GRPO}}^{(2)}(\theta)
    + \alpha(t)\,\mathcal{L}_{\mathrm{scorer}}(\theta).
    \label{eq:stage2_loss}
\end{equation}

Here, $\alpha(t)$ is a scheduled coefficient that activates the scorer calibration loss after the initial generator warm-up, improving training stability under the shared parameters. Under this joint objective, the generator learns query-adaptive rubrics beyond the static seed rubrics. The tips $T_k$ can adapt to prompt-specific details, while the weights $w_k$ vary with the query; meanwhile, the scorer remains guided by the human-aligned margin loss. A qualitative example is provided in Figure~\ref{fig3}.

\section{Experimental Evaluation}
\label{sec:experiments}

We evaluate RewardVerse from four aspects: (i) pointwise correlation with human ratings across the 16 EvalVerse dimensions; (ii) transfer to pairwise ranking on VGRB; (iii) the contribution of each RGPO component; and (iv) downstream use as an RL reward for video generation.

\subsection{Benchmarks and Baselines}
\label{sec:benchmarks}

We validate our method on two complementary benchmarks. First, we curate a pointwise test set from \textbf{EvalVerse}~\citep{yang2026evalverse}, currently the most comprehensive evaluation suite for generative video models, spanning 16 fine-grained secondary dimensions (details in Appendix~\ref{sec:appendix_dim}). Second, we evaluate on \textbf{VideoGen-RewardBench (VGRB)}~\citep{liu2026improving}, a large-scale external pairwise preference benchmark of over $26.5$K video pairs. We report the Visual Quality split as transfer on a seen evaluation dimension and the Text Alignment split as transfer to an unseen dimension.

We train our method on Qwen2.5-VL-7B~\citep{qwen25vl} and compare against six state-of-the-art video reward models: VideoScore-v1.1~\citep{he2024videoscore}, VideoScore2~\citep{he2025videoscore2}, UnifiedReward and its thinking variant~\citep{unifiedreward}, VideoReward~\citep{liu2026improving}, VisionReward~\citep{xu2026visionreward}, and Q-Scorer~\citep{tang2026revisiting}. To trace the effect of each stage in our pipeline, we further analyze two internal baselines: \textbf{Raw-seed Rubric} (zero-shot with the seed rubric), \textbf{Raw-rubric} (zero-shot with dynamically generated but uncalibrated rubrics). 

\begin{table}[t]
\vspace{-0.4cm}
\caption{Per-dimension \textbf{PLCC} (top row) and \textbf{SRCC} (bottom row) on the 16 secondary dimensions of EvalVerse (dimension names defined in Appendix~\ref{sec:appendix_dim}). \textbf{Avg.} denotes the macro-average over the 10 dimensions supported by all compared reward models. Best in \textbf{bold}, second-best \underline{underlined}. `\textendash' means the reward model does not natively support that dimension.}
\label{tab:pointwise}
\centering
\scriptsize
\setlength{\tabcolsep}{2.0pt}
\resizebox{\textwidth}{!}{
\begin{tabular}{ll ccccccccccccccccc}
\toprule
\textbf{Method} &
& \textbf{Scene} & \textbf{Const.} & \text{Action} & \textbf{Expr.}
& \textbf{Comp.} & \textbf{Lens} & \textbf{Pace} & \textbf{Visual}
& \textbf{Chrom.} & \textbf{Mater.} & \textbf{Light.} & \textbf{Ground.}
& \textbf{Prog.} & \textbf{Logic} & \textbf{Rhythm} & \textbf{Char.}
& \textbf{Avg.} \\
\midrule
\multicolumn{19}{l}{\emph{External Reward Models}} \\
\multirow{2}{*}{VideoScore-v1.1}
 & PLCC & -0.086 & 0.102 & -0.038 & \textendash & \textendash & \textendash & \textendash & 0.119 & -0.105 & -0.156 & -0.031 & \textendash & \textendash & -0.245 & 0.070 & 0.042 & -0.033 \\
 & SRCC & 0.002 & 0.255 & -0.042 & \textendash & \textendash & \textendash & \textendash & 0.242 & 0.117 & 0.012 & 0.195 & \textendash & \textendash & 0.099 & 0.280 & 0.155 & 0.132 \\
\multirow{2}{*}{VideoScore2}
 & PLCC & 0.449 & 0.245 & \underline{0.390} & \textendash & \textendash & \textendash & \textendash & 0.507 & \underline{0.367} & \textbf{0.697} & 0.534 & \textendash & \textendash & 0.372 & 0.427 & 0.346 & 0.433 \\
 & SRCC & 0.443 & 0.337 & \underline{0.363} & \textendash & \textendash & \textendash & \textendash & 0.481 & \textbf{0.324} & \textbf{0.698} & \underline{0.502} & \textendash & \textendash & 0.370 & 0.340 & 0.409 & \underline{0.427} \\
\multirow{2}{*}{UnifiedReward}
 & PLCC & 0.461 & 0.307 & 0.175 & \textendash & \textendash & \textendash & \textendash & 0.404 & -0.020 & 0.333 & 0.463 & \textendash & \textendash & 0.465 & 0.359 & 0.396 & 0.334 \\
 & SRCC & \underline{0.468} & 0.321 & 0.183 & \textendash & \textendash & \textendash & \textendash & 0.332 & 0.063 & 0.317 & 0.340 & \textendash & \textendash & \underline{0.549} & 0.354 & 0.355 & 0.328 \\
\multirow{2}{*}{VideoReward}
 & PLCC & 0.313 & \underline{0.381} & 0.326 & \textendash & \textendash & \textendash & \textendash & 0.462 & 0.237 & 0.573 & 0.483 & \textendash & \textendash & 0.326 & 0.265 & 0.474 & 0.384 \\
 & SRCC & 0.334 & 0.360 & 0.354 & \textendash & \textendash & \textendash & \textendash & 0.556 & 0.220 & \underline{0.622} & 0.471 & \textendash & \textendash & 0.334 & 0.304 & 0.430 & 0.399 \\
\multirow{2}{*}{VisionReward}
 & PLCC & 0.457 & 0.246 & 0.332 & \textendash & \textendash & \textendash & \textendash & 0.554 & 0.234 & 0.524 & 0.514 & \textendash & \textendash & 0.399 & 0.387 & 0.429 & 0.408 \\
 & SRCC & 0.400 & 0.220 & 0.186 & \textendash & \textendash & \textendash & \textendash & 0.519 & 0.191 & 0.508 & 0.479 & \textendash & \textendash & 0.171 & \underline{0.370} & 0.466 & 0.351 \\
\multirow{2}{*}{Q-Scorer}
 & PLCC & \underline{0.496} & 0.356 & 0.182 & \textendash & \textendash & \textendash & \textendash & \underline{0.517} & 0.234 & \underline{0.614} & \textbf{0.623} & \textendash & \textendash & \underline{0.593} & \underline{0.517} & \underline{0.534} & \underline{0.467} \\
 & SRCC & 0.340 & 0.277 & 0.139 & \textendash & \textendash & \textendash & \textendash & 0.370 & 0.230 & 0.505 & 0.451 & \textendash & \textendash & 0.337 & \textbf{0.417} & \underline{0.476} & 0.354 \\
\midrule
\multicolumn{19}{l}{\emph{RGPO Pipeline Stages}} \\
\multirow{2}{*}{Raw-seed Rubric}
 & PLCC & 0.432 & 0.227 & 0.144 & 0.290 & \underline{0.547} & 0.303 & \underline{0.079} & 0.236 & 0.120 & 0.523 & 0.173 & \underline{0.441} & \underline{0.225} & 0.589 & 0.194 & 0.471 & 0.311 \\
 & SRCC & 0.449 & \underline{0.372} & 0.141 & 0.251 & \underline{0.275} & \underline{0.278} & \underline{0.159} & 0.266 & 0.148 & 0.404 & 0.213 & \underline{0.341} & \underline{0.251} & 0.478 & -0.043 & 0.326 & 0.275 \\
\multirow{2}{*}{Raw-rubric}
 & PLCC & 0.437 & 0.175 & 0.277 & \underline{0.415} & 0.449 & \underline{0.484} & 0.075 & \underline{0.571} & 0.152 & 0.531 & 0.373 & 0.061 & 0.183 & 0.583 & 0.247 & 0.447 & 0.379 \\
 & SRCC & 0.417 & 0.178 & 0.293 & \textbf{0.340} & 0.237 & 0.166 & 0.065 & \underline{0.557} & 0.055 & 0.457 & 0.309 & 0.034 & 0.213 & 0.297 & 0.107 & 0.263 & 0.293 \\
\multirow{2}{*}{\textbf{Joint (Ours)}}
 & PLCC & \textbf{0.520} & \textbf{0.395} & \textbf{0.566} & \textbf{0.448} & \textbf{0.575} & \textbf{0.486} & \textbf{0.471} & \textbf{0.661} & \textbf{0.498} & 0.538 & \underline{0.557} & \textbf{0.456} & \textbf{0.510} & \textbf{0.750} & \textbf{0.523} & \textbf{0.535} & \textbf{0.554} \\
 & SRCC & \textbf{0.493} & \textbf{0.381} & \textbf{0.403} & \underline{0.285} & \textbf{0.440} & \textbf{0.333} & \textbf{0.431} & \textbf{0.610} & \underline{0.243} & 0.514 & \textbf{0.515} & \textbf{0.391} & \textbf{0.324} & \textbf{0.656} & 0.146 & \textbf{0.500} & \textbf{0.446} \\
\bottomrule
\end{tabular}
}
\vspace{-0.5cm}
\end{table}

\subsection{Pointwise Evaluation on EvalVerse}
\label{sec:pointwise-main}

Table~\ref{tab:pointwise} reports the per-dimension pointwise evaluation results across the 16 EvalVerse dimensions. RewardVerse (\textbf{Joint (Ours)}) achieves the strongest overall performance across the 16 dimensions, with the highest PLCC in 14 dimensions. It outperforms other models by a clear margin on challenging cognitive and temporal tasks, such as Logic (PLCC of $0.750$ vs. the next-best $0.593$) and Action (PLCC of $0.566$ vs. $0.390$). These results show that RGPO effectively aligns dynamic rubric-guided scores with human judgments.
External video scorers can be poorly calibrated on these fine-grained axes (e.g., VideoScore-v1.1 has a PLCC of $-0.245$ on Logic). Interestingly, Q-Scorer remains competitive on some non-quality dimensions such as Lighting and Logic, suggesting that these dimensions are partly correlated with low-level visual quality. Beyond correlation, RewardVerse also shows better score alignment with human ratings after RGPO, as further analyzed in Appendix~\ref{app:post_rgpo_drift}.

\subsection{Pairwise Transfer on VGRB}
\label{sec:pairwise}

Table~\ref{tab:vgrb} reports the pairwise agreement on the large-scale external benchmark VGRB. Most notably, RewardVerse achieves the best performance among all non-oracle baselines on both splits. Despite training on only 30 pairs per dimension (480 pairs in total), our model outperforms the strongest non-oracle baseline (VisionReward) on the seen Visual Quality (VQ) split by $7.1$ (Acc w/ Tie) and $7.0$ (Acc w/o Tie) percentage points, closing much of the gap to the oracle VideoReward.
Furthermore, our model demonstrates strong zero-shot generalization on the unseen Text Alignment (TA) split. It achieves the highest accuracy ($0.471$ with ties and $0.623$ without ties). These results show that mitigating scalar drift in pointwise scoring can also improve pairwise preference alignment, even when transferring to out-of-distribution benchmarks and unseen evaluation dimensions.

\begin{table}[t]
\vspace{-0.4cm}
\caption{Pairwise agreement on VGRB. VQ is seen for RewardVerse; TA is unseen at training. VideoReward is the model whose training set defines VGRB and is shown in gray as an oracle upper bound rather than a comparable baseline.}
\label{tab:vgrb}
\centering
\small
\begin{tabular}{lcccc}
\toprule
& \multicolumn{2}{c}{\textbf{Text Alignment (unseen)}} &
  \multicolumn{2}{c}{\textbf{Visual Quality (seen)}} \\
\cmidrule(lr){2-3}\cmidrule(lr){4-5}
\textbf{Method} & Acc w/~Tie & Acc w/o~Tie & Acc w/~Tie & Acc w/o~Tie \\
\midrule
VideoScore-v1.1        & 0.372 & 0.503 & 0.474 & 0.477 \\
VideoScore2          & 0.301 & 0.497 & 0.374 & 0.630 \\
UnifiedReward          & 0.303 & 0.564 & 0.412 & 0.394 \\
UnifiedReward-Thinking & 0.428 & 0.582 & \textendash & \textendash \\
VisionReward           & 0.465 & 0.611 & 0.474 & 0.590 \\
\textbf{RewardVerse (ours)}
                       & \textbf{0.471} & \textbf{0.623} & \textbf{0.545} & \textbf{0.660} \\
\midrule
\textcolor{gray}{VideoReward (oracle upper bound)}
                       & \textcolor{gray}{0.538} & \textcolor{gray}{0.722}
                       & \textcolor{gray}{0.596} & \textcolor{gray}{0.756} \\
\bottomrule
\end{tabular}
\vspace{-0.5cm}
\end{table}

\subsection{Ablation Studies and Downstream Evaluation}
\label{sec:ablation}

\begin{table}[t]
\caption{Component-wise ablation on EvalVerse (averaged over 16 dimensions). Best in \textbf{bold}.}
\label{tab:ablation}
\centering
\small
\begin{tabular}{lcc}
\toprule
\textbf{Configuration} & \textbf{PLCC}~$\uparrow$ & \textbf{SRCC}~$\uparrow$ \\
\midrule
\textbf{Full RGPO}                                & \textbf{0.530} & \textbf{0.416} \\
% \quad (a) w/o Calibration loss $\mathcal{L}_{\mathrm{cal}}$ & \textendash & \textendash & \textendash \\
\quad (a) w/o Stage-1 Scorer Warm-up                        & 0.431 & 0.355 \\
\quad (b) w/o Stage-2 Joint Optimization                        & 0.456 & 0.364 \\
\quad (c) w/o Hierarchical tips $T_k$ (flat rubric)         & 0.497 & 0.398 \\
\midrule
\multicolumn{3}{l}{\emph{Rubric-Free Baselines (equal $480$-pair budget)}} \\
\quad (d) V2-Raw (zero-shot, soft-logits)     & 0.361 & 0.234 \\
\quad (e) V2-Training (direct pointwise training) & 0.439 & 0.351 \\
\bottomrule
\end{tabular}
\vspace{-0.5cm}
\end{table}

Table~\ref{tab:ablation} reports macro-averaged PLCC and SRCC on EvalVerse, together with two rubric-free baselines under the same 480-pair budget. Three findings emerge. First, using a rubric as an intermediate representation is important for low-data efficiency. Under the same 480-pair budget, direct rubric-free pointwise training (e) improves only slightly over its zero-shot counterpart (d) and remains well below Full RGPO, showing the benefit of introducing the rubric between the query and the scorer when training data is limited. Second, Stage-1 scorer warm-up is important for score calibration. Removing Stage 1 (a) drops PLCC by 0.099 and SRCC by 0.061, showing the benefit of calibrating the scorer before dynamic rubric learning begins. Third, Stage-2 joint optimization improves over static rubrics. Skipping Stage 2 (b) yields a 0.074 PLCC gap compared with the full model, confirming the benefit of adapting rubrics to different queries. Qualitative results are provided in Appendix~\ref{app:qualitative}.

\paragraph{Downstream Video Generation.}
We further use RewardVerse as the reward for GRPO fine-tuning of Wan-2.2-A14B~\citep{wan2025wan} on Visual Quality. RewardVerse improves Imaging Quality from 0.640 to 0.653, while VBench-Quality remains stable (0.808 to 0.809) and VBench-Text improves from 0.428 to 0.446. In comparison, VideoReward improves Imaging Quality to 0.648 but reduces VBench-Quality and VBench-Text to 0.804 and 0.392. These results show that RewardVerse improves the target quality while better preserving other video capabilities. Full settings and results are provided in Appendix~\ref{app:downstream}.

\section{Conclusion}
\label{sec:conclusion}
In this paper, we introduced RewardVerse, a robust pointwise video reward model, and its training algorithm, Rubric-Guided Policy Optimization (RGPO). Addressing scalar drift in unconstrained MLLM scoring, we demonstrated that inserting a dynamic rubric improves score resolution, while RGPO further alleviates the remaining scalar drift through joint optimization. We also showed that both stages of RGPO—seed-guided warm-up and joint optimization—are essential to this process. With only 30 preference pairs per dimension, RewardVerse achieves high pointwise correlation on EvalVerse, strong pairwise agreement on VGRB, and improves downstream video generation while being less prone to reward hacking. These results show that mitigating scalar drift is crucial for using video reward models as reinforcement learning signals.

\bibliography{iclr2027_conference}
\bibliographystyle{iclr2027_conference}

\newpage
\appendix
\section{Appendix}

\textbf{Code, models, and project page:} \url{https://github.com/2kxx/RewardVerse}.

\subsection{Extended Empirical Diagnostics on Scalar Drift}
\label{app:scalar_drift}

\subsubsection{Full 4-variant Ablation for Qwen2.5-VL}
\label{app:full-ablation}
Table~\ref{tab:scalar_drift_main} expands upon the core conclusions of Section~\ref{sec:variants}. To ensure a controlled and rigorous evaluation, all four variants are evaluated using the same multi-modal LLM backbone (Qwen2.5-VL-7B). They share an identical pointwise hold-out set ($n_{\mathrm{pw}} = 240$) and pairwise comparison pool ($n_{\mathrm{pair}} = 240$), spanning sixteen key dimensions from EvalVerse's global taxonomy to ensure comprehensive coverage. All inference hyperparameters are held constant across variants, utilizing a frame rate of $\text{FPS} = 2$ and greedy decoding. For evaluation metrics, we let $\sigma(\hat{s})$ denote the standard deviation of predicted scores, $\text{bias} = \overline{\hat{s}} - \overline{s_{\text{label}}}$ represent the systematic overestimation error, and $\overline{|\Delta|}$ represent the mean prediction gap on decided pairs (i.e., those whose constituent videos receive distinct scores).

\begin{table}[h]
\vspace{-0.5cm}
\centering
\caption{A four-variant ablation study on the white-box \textbf{Qwen2.5-VL-7B} backbone across the shared 16-dimensional hold-out set ($n_{\mathrm{pw}} = 240$) and pair pool ($n_{\mathrm{pair}} = 240$). The deployed configuration (V3) is highlighted in \textbf{bold}.}
\label{tab:scalar_drift_main}
\setlength{\tabcolsep}{4.5pt}
\small
\begin{tabular}{lccrrrrrrr}
\toprule
Variant & Rubric & Format
        & $\overline{\hat{s}}$
        & $\sigma(\hat{s})$
        & Bias
        & Decided
        & Pair Acc.
        & $\overline{|\Delta|}$ \\
\midrule
V1 (NL free-float)               & \ding{55} & natural    & 4.747 & 0.203 & +1.097 & 58/240   & $0.931^{\dag}$ & 0.338 \\
V2 (soft-logits)                  & \ding{55} & soft-log.  & 4.194 & 0.258 & +0.546 & 239/240  & $0.795$        & 0.168 \\
\textbf{V3 (rubric + soft-logits)} & \ding{51} & soft-log.  & 4.419 & \textbf{0.471} & +0.771 & \textbf{240/240} & $0.754$ & \textbf{0.216} \\
V4 (rubric + multi-theme NL float) & \ding{51} & natural    & 4.649 & 0.275 & +1.001 & 76/240   & $0.737^{\dag}$ & 0.425 \\
\bottomrule
\end{tabular}\\[2pt]
{\footnotesize $^{\dag}$~V1 and V4 produce only a small portion of distinct pair scores (the rest tie); the reported pair accuracy is evaluated on the decided subset and therefore over-states real-world separability.}
\vspace{-0.5cm}
\end{table}

\subsubsection{Large-Scale Evaluation of Gemini-3.1-Pro on 1K Test Set}
\label{app:gemini-1k}
\begin{figure}[h]
\centering
\includegraphics[width=0.83\linewidth]{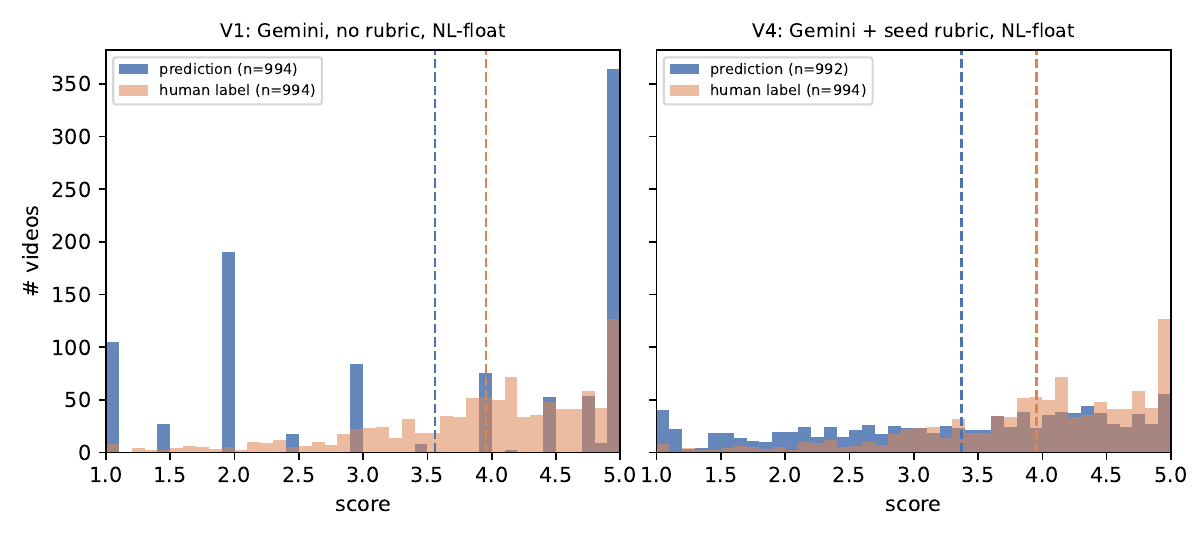}
\vspace{-0.5cm}
\caption{Pointwise prediction histograms of Gemini-3.1-Pro (V1 vs. V4) on the complete 1K pointwise test set, overlaid on human gold standard ratings (orange). Rubrics improve correlation but worsen calibration. Decomposing evaluations (V4) aligns the rating variance closer to human baselines, reducing the uncalibrated standard deviation from 1.500 (V1) to 1.158 (V4).}
\label{fig:scalar_drift_hist_gemini}
\vspace{-0.5cm}
\end{figure}

To evaluate the scalability and generalizability of our rubric-guided paradigm on frontier models, we extend our evaluation to \textbf{Gemini-3.1-Pro} using the complete 1K multi-dimensional pointwise test set. Given that closed-source APIs restrict access to token-level logits, we benchmark the V1 (No Rubric) and V4 (With Rubric) configurations under natural-language float output formats. 
Table~\ref{tab:scalar_drift_gemini} lists the PLCC, SRCC, standard deviation and mean scores on the 1K dataset, while Figure~\ref{fig:scalar_drift_hist_gemini} shows the distribution of our predictions against human labels.

\begin{table}[h]
\vspace{-0.5cm}
\centering
\caption{Comprehensive pointwise calibration and correlation metrics of \textbf{Gemini-3.1-Pro} on the large-scale \textbf{1K pointwise test set} ($16$ evaluation dimensions) under V1 and V4 configurations. Standard deviation, biases, and discrete rating distributions are included to show scaling behavior and systematic offsets.}
\label{tab:scalar_drift_gemini}
\setlength{\tabcolsep}{5pt}
\small
\begin{tabular}{lcccccccc}
\toprule
Variant & Rubric & $\overline{\hat{s}}$ & $\sigma(\hat{s})$ & Bias & $\mathrm{PLCC}$ & $\mathrm{SRCC}$ \\
\midrule
V1 & \ding{55} & 3.558 & 1.500 & -0.398 & +0.490 & +0.480 \\
V4 & \ding{51} & 3.372 & 1.158 & -0.584 & \textbf{+0.593} & \textbf{+0.564} \\
\bottomrule
\end{tabular}
\end{table}

The empirical results on the 1K dataset demonstrate that the rubric-as-an-intermediate-representation paradigm generalizes robustly to proprietary, state-of-the-art systems. Integrating the rubric ($V1 \to V4$) unlocks substantial alignment gains: Pearson correlation (PLCC) increases by +0.103 (a 21.0\% relative improvement), while Spearman rank correlation (SRCC) increases by +0.084 (17.5\% relative), demonstrating that rubric-guided sub-dimensions align much closer with human multi-dimensional preferences.

Notably, the calibration behavior of Gemini-3.1-Pro reveals a striking contrast to Qwen2.5-VL. While Qwen2.5-VL suffers from systematic overestimation (positive bias), Gemini-3.1-Pro exhibits a persistent underestimation (negative bias of $-0.398$ in V1 and $-0.584$ in V4).
This divergence demonstrates that different foundational models, regardless of their scale, possess intrinsic score-scale biases. Relying solely on zero-shot static rubrics cannot eliminate these systematic biases, which strongly justifies the necessity of our proposed RGPO training paradigm to explicitly align the reward model with human anchors.

\subsubsection{Scalar Drift across Existing Reward Models}
\label{app:cross_rm_drift}

We further examine the score distributions of six existing reward models on the shared 10-dimensional test subset. Since these models use different native output ranges, we remap their predictions to $[1,5]$ for visualization. This remapping is applied only to display the distributions.

\begin{figure}[t]
    \centering
    \includegraphics[width=0.9\textwidth]{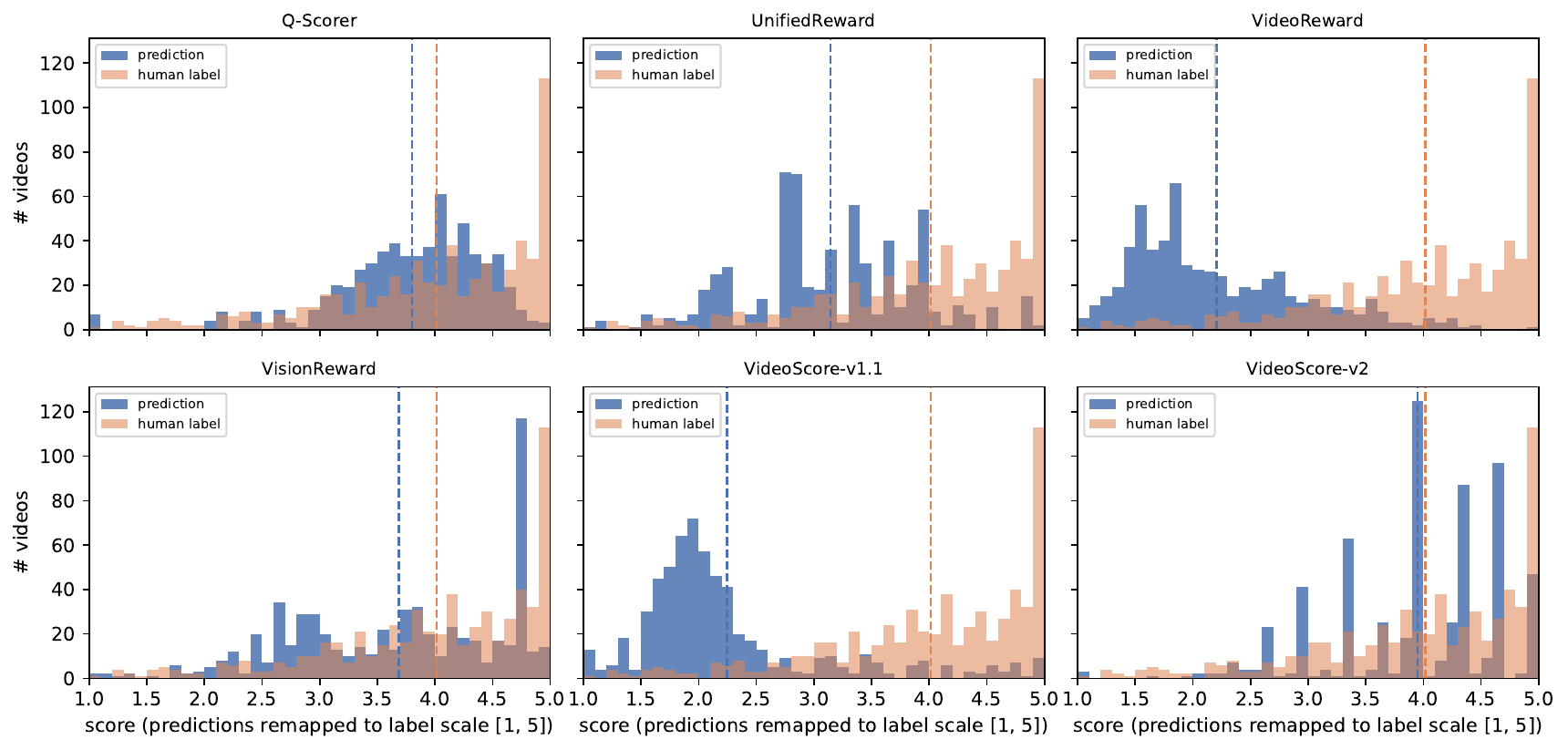}
    \caption{
    Score distributions of six reward models on the shared 10-dimensional test subset. Predictions are remapped to the human rating scale $[1,5]$ for visualization. Blue bars denote model predictions and orange bars denote human labels. Dashed lines show their means. Most reward models exhibit score-range compression or distribution shift. Q-Scorer produces a relatively broader score distribution, but it still does not fully match human ratings.
    }
    \label{fig:cross_rm_scalar_drift}
    \vspace{-0.5cm}
\end{figure}

As shown in Figure~\ref{fig:cross_rm_scalar_drift}, most reward models produce scores within a narrow range or show a clear shift from the human rating distribution. VideoReward and VideoScore-v1.1 mainly concentrate in the low-score region, while VisionReward and VideoScore-v2 place many predictions near the upper end. UnifiedReward also produces several narrow score peaks. Q-Scorer is the only model that gives a relatively broad and continuous score distribution, although its mean and overall distribution still differ from human ratings. These results show that score compression or scale shift is common across existing reward models.

\subsubsection{Per-Dim Breakdown}
\label{app:calibration-wall}
\begin{figure}[h]
\centering
\includegraphics[width=0.78\linewidth]{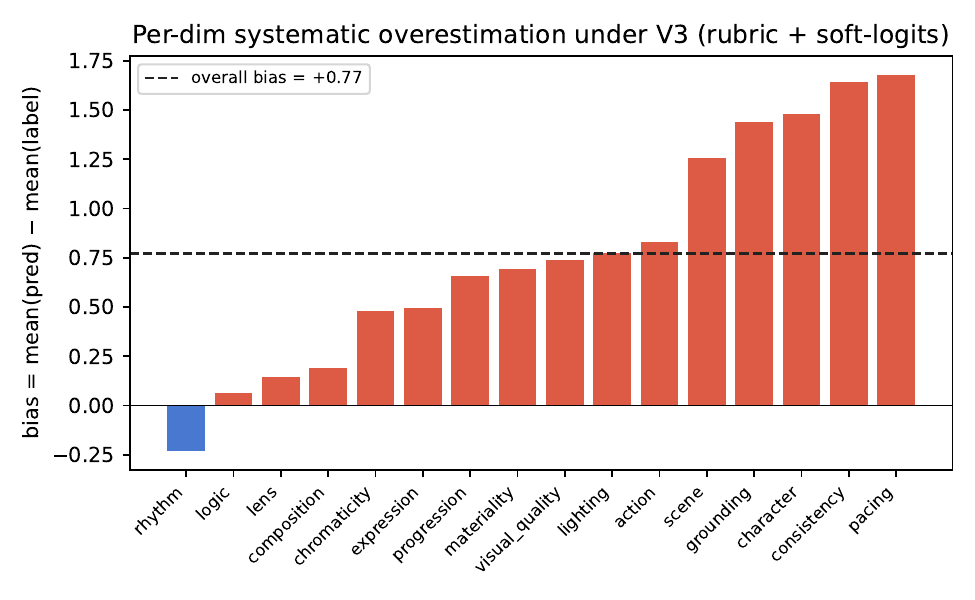}
\vspace{-0.5cm}
\caption{Per-dimension systematic overestimation bias of the V3 protocol ($n = 15$ per dimension). Bars represent the difference between predicted means and human ratings ($\overline{\hat{s}_d} - \overline{s_{\text{label},d}}$), with the dashed line marking the pooled bias ($+0.77$). The pervasive positive offset across 15 out of 16 dimensions highlights a limitation that static, zero-shot rubric prompting fails to resolve.}
\label{fig:scalar_drift_per_dim_bias}
\vspace{-0.5cm}
\end{figure}

The pooled bias of the deployed V3 configuration is $+0.771$ (bootstrap $95\%$ CI: $[+0.64, +0.90]$, $n_{\mathrm{boot}} = 2000$, $n = 240$). Variants V1 ($+1.10$), V2 ($+0.55$), and V4 ($+1.00$) similarly overshoot human ratings, ruling out the rubric scaffold as the primary source of overestimation. As shown in Figure~\ref{fig:scalar_drift_per_dim_bias}, 15 out of 16 evaluation dimensions exhibit positive bias—with \textit{rhythm} being the sole negative exception ($-0.23$). The largest calibration gaps occur in \textit{pacing} ($+1.68$), \textit{consistency} ($+1.64$), and \textit{character} ($+1.48$). This systematic positive offset constitutes the limitation that our proposed RGPO training algorithm (Section~\ref{sec:method}) aims to mitigate.

A dimensional analysis reveals a clear semantic divide: higher-level temporal reasoning and consistency dimensions (e.g., \textit{pacing}, \textit{consistency}) suffer from severe overestimation, whereas low-level sensory dimensions (e.g., \textit{rhythm}) show minimal bias. This pattern suggests that uncalibrated MLLMs act as overly optimistic critics: they align well with low-level sensory cues but remain blind to fine-grained temporal artifacts and physical plausibility. An explicit reinforcement learning loop (RGPO) is vital to break through this limitation.

\subsubsection{Protocol Selection: How to Use the Rubric}
\label{app:protocols}
Section~\ref{sec:variants} establishes that the deployed evaluator must combine a rubric with a soft-logits readout. A secondary design question remains: \emph{within} the rubric paradigm, what is the best way to query and format the prompt? We benchmark four candidate protocols on the \textit{visual quality} hold-out set ($n_{\mathrm{pw}} = 30$, $n_{\mathrm{pair}} = 30$), all utilizing the same rubric:
\begin{itemize}
    \item \textbf{PR1 (Pairwise A/B):} The model evaluates both videos simultaneously in a single forward pass, selecting video A or B with the rubric injected as an evaluation reference.
    \item \textbf{PR2 (Pointwise Overall Integer):} One forward pass is conducted per video; the model is prompted to emit a single integer score in the range $[1, 5]$.
    \item \textbf{PR3 (Pointwise Multi-Theme Integer):} One forward pass is conducted per video; the model outputs a JSON object containing per-theme integer scores.
    \item \textbf{PR4 (Pointwise Multi-Theme Soft-Logits):} One forward pass is conducted per video; the continuous soft-logits expectation is computed for each rubric theme, and the overall score is aggregated via a weighted average. This is the deployed configuration (V3).
\end{itemize}
The \emph{format-failure rate} counts forward passes that either cannot be successfully parsed as valid JSON/integers. Ties at integer granularity are reflected separately by the number of decided pairs. Since PR1 directly outputs a binary preference rather than a scalar value, its correlation metrics (PLCC, SRCC) and score margins are not defined.

\begin{table}[h]
\vspace{-0.3cm}
\centering
\caption{A comparative evaluation of different rubric-querying protocols on the \textit{visual\_quality} hold-out set ($30$ pointwise samples and $30$ evaluation pairs). The proposed \textbf{PR4} is the unique protocol that simultaneously achieves a $0\%$ format-failure rate, generates continuous scalar outputs, and preserves full pairwise separability ($100\%$ decided pairs).}
\label{tab:protocols}
\begin{tabular}{lccccc}
\toprule
Protocol & Format-Fail Rate & Decided & Pair Acc. & PLCC & $\overline{|\Delta|}$ \\
\midrule
PR1 --- pairwise A/B                    & 0.000          & 30/30  & 1.000 & ---   & ---   \\
PR2 --- pointwise integer (text)        & 0.467          & 9/30   & 1.000 & 0.313 & 1.333 \\
PR3 --- multi-theme integer (text)      & 0.533          & 6/30   & 0.833 & 0.374 & 1.667 \\
\textbf{PR4 --- multi-theme soft-logits}& \textbf{0.000} & \textbf{30/30}& 0.833 & 0.229 & 0.269 \\
\bottomrule
\end{tabular}
\vspace{-0.5cm}
\end{table}

Although PR1 achieves perfect pairwise accuracy, it is unsuitable as an RL reward signal due to its lack of scalar outputs. Furthermore, PR1 is highly sensitive to input ordering (see Appendix~\ref{app:perturbation}), consistent with contextual positional bias observed in video-language models \citep{xia2026video}. While proprietary models (e.g., Gemini-3.1-Pro) handle multi-video comparison robustly, using them as iterative RL reward models is computationally and financially prohibitive.

For integer-based configurations (PR2 and PR3), formatting parser failures and scale quantization ties severely hinder evaluation, yielding failure rates of 46.7\% and 53.3\%, respectively. These high rates demonstrate that free-form text score generation is fragile, as minor JSON variations or tokenization shifts block automated parsing.
Conversely, PR4 bypasses parsing bottlenecks by extracting soft logits directly from vocabulary tokens, guaranteeing a 0\% failure rate by construction. By avoiding multi-video reasoning bottlenecks and providing a continuous optimization surface, PR4 serves as our final deployment variant.

\subsubsection{Stability under Prompt Perturbation and Pair Swap}
\label{app:perturbation}
Two diagnostic studies probe the second symptom of scalar drift: \emph{high variance under context shift}.

\paragraph{Diagnostic 1: Prompt perturbation (V2 vs. V3).}
We evaluate 48 videos (3 per dimension across 16 dimensions) three times under three semantically equivalent overall-quality instructions, reporting the mean per-video standard deviation. We compare V2 and V3 because both utilize the soft-logits readout, leaving the rubric as the sole independent variable. Variant V1 is excluded as its scores collapse onto a single peak, rendering stability metrics meaningless. To ensure commensurability between the two distinct scoring scales, each variant's per-video standard deviation is normalized by its global standard deviation $\sigma(\hat{s})$ from Table~\ref{tab:scalar_drift_main}.

\begin{table}[h]
\vspace{-0.3cm}
\centering
\caption{Per-video score standard deviation across $K=3$ paraphrased instructions ($N=48$ videos). Statistics are computed over per-video prompt-perturbation standard deviations $\{\sigma_i\}_{i=1}^N$. ``Normalized'' reports the Raw Mean scaled by the global standard deviation $\sigma(\hat{s})$.}
\label{tab:b1_perturb}
\begin{tabular}{lcccc}
\toprule
                            & Raw Mean & Median & Max & Normalized \\
\midrule
V2 (soft-logits, no rubric) & 0.046 & 0.040 & 0.120 & 0.179 \\
V3 (rubric + soft-logits)   & 0.094 & 0.038 & 0.418 & 0.199 \\
\bottomrule
\end{tabular}
\vspace{-0.3cm}
\end{table}

After normalization, both variants exhibit comparable relative variance under prompt paraphrase (0.18 vs. 0.20). This demonstrates a crucial structural benefit: adding the rubric (V3) decompresses the scoring scale by 82.6\% ($\sigma(\hat{s})$ increases from 0.258 to 0.471) while maintaining similar relative stability under prompt paraphrases. The smaller raw variation of V2 mainly results from its compressed score range. Under the soft-logits readout, the rubric substantially expands score resolution without causing a comparable increase in relative sensitivity to prompt paraphrases.

\paragraph{Diagnostic 2: Pair A/B order swap (no rubric vs. seed rubric).}
We evaluate every video pair in both input orderings, $(y_w,\,y_l)$ and $(y_l,\,y_w)$, and report the rate at which preference verdicts disagree across the swap (flip rate). The pairwise A/B protocol is tested on 48 sampled pairs (3 per dimension) under two settings: \emph{without} the rubric (baseline) and \emph{with} the seed per-dimension rubric injected as a reference (PR1). Table~\ref{tab:b2_swap} summarizes the results.

\begin{table}[h]
\vspace{-0.3cm}
\centering
\caption{Pair A/B swap flip rates with and without the rubric ($48$ pairs). Injecting the rubric substantially reduces the flip rate, yet the residual bias remains a significant bottleneck for the pairwise protocol.}
\label{tab:b2_swap}
\begin{tabular}{lcc}
\toprule
Setting                       & Decided pairs & Flip rate \\
\midrule
Pairwise A/B, no rubric       & 48/48         & 0.792     \\
Pairwise A/B, seed rubric     & 48/48         & 0.562     \\
\bottomrule
\end{tabular}
\vspace{-0.5cm}
\end{table}

Introducing the rubric reduces the flip rate by 23.0 percentage points (from 0.792 to 0.562), confirming that a rubric serves as a grounding reference that aligns model decisions with objective criteria rather than order-sensitive heuristics. However, a flip rate of 56.2\% remains very high, confirming that the binary A/B format is highly vulnerable to position bias when processing two complex video sequences simultaneously.

These findings yield two critical conclusions. First, this high residual flip rate disqualifies the pairwise A/B protocol from serving as an RL reward model, as unstable preference signals would generate highly noisy gradients. This provides an independent justification for adopting our pointwise multi-theme soft-logits protocol (PR4 / V3). Second, it highlights the division of labor in our framework: the rubric calibrates the scoring criteria, while the pointwise soft-logits format ensures greater robustness under context shifts by completely bypassing order-dependent comparison bottlenecks.

\subsubsection{Scalar Drift Before and After RGPO}
\label{app:post_rgpo_drift}

The four-variant study in Appendix~\ref{app:full-ablation} analyzes the effect of rubric injection and score readout before training. We further evaluate whether the complete RGPO training further mitigates the remaining scalar drift. We compare the original Qwen2.5-VL-7B model with the jointly trained RewardVerse model. Both models generate a dynamic rubric for each query and use the same multi-theme soft-logits scoring protocol. Score calibration is evaluated on the same pointwise hold-out set ($n=240$) in Appendix~\ref{app:full-ablation}, while prompt stability is evaluated on the 48 videos and three paraphrased instructions used in Appendix~\ref{app:perturbation}.

\begin{table}[t]
\centering
\caption{Pointwise scoring before and after RGPO under the same dynamic-rubric protocol. Bias is the difference between the mean prediction and the mean human rating. The 95\% confidence interval is obtained by bootstrap sampling.}
\label{tab:post_rgpo_drift}
\begin{tabular}{lccccccc}
\toprule
Model & $N$ & Mean & $\sigma(\hat{s})$ & Bias & 95\% CI & PLCC & SRCC \\
\midrule
Pre-RGPO  & 240 & 4.198 & 0.457 & +0.610 &
[+0.488, +0.736] & 0.240 & 0.222 \\
Post-RGPO & 240 & 3.752 & 1.332 & +0.164 &
[+0.012, +0.313] & 0.505 & 0.458 \\
\bottomrule
\end{tabular}
\vspace{-0.5cm}
\end{table}

As shown in Table~\ref{tab:post_rgpo_drift}, the untrained model remains systematically optimistic even when using dynamic rubrics, with a mean bias of $+0.610$. Although RGPO is optimized for relative score margins rather than absolute scores, the bias decreases to $+0.164$, while PLCC and SRCC increase from $0.240$ and $0.222$ to $0.505$ and $0.458$, respectively. The prediction standard deviation also increases from $0.457$ to $1.332$, showing that RGPO further expands the previously compressed score range. At the dimension level, the absolute calibration bias is reduced on 10 out of 16 dimensions, with the largest reductions appearing in pacing, materiality, character, lighting, and scene. We additionally repeat the prompt-perturbation diagnostic under the same dynamic-rubric protocol. For each video, we compute the standard deviation of its scores across three semantically equivalent instructions. Since RGPO substantially expands the overall score range, we normalize the mean
per-video standard deviation by the global prediction standard deviation.

\begin{table}[t]
\centering
\caption{
Prompt-perturbation stability before and after RGPO under the same dynamic-rubric scoring protocol. Raw statistics are computed over the per-video standard deviations across three paraphrased instructions. Normalized is the Raw Mean divided by the global prediction standard
deviation.
}
\label{tab:rgpo_prompt_stability}
\begin{tabular}{lccccc}
\toprule
Model & Raw Mean & Median & Max & Global $\sigma(\hat{s})$ & Normalized \\
\midrule
Pre-RGPO  & 0.094 & 0.038 & 0.418 & 0.457 & 0.206 \\
Post-RGPO & 0.113 & 0.042 & 0.755 & 1.332 & \textbf{0.085} \\
\bottomrule
\end{tabular}
\vspace{-0.5cm}
\end{table}

As shown in Table~\ref{tab:rgpo_prompt_stability}, the raw variation slightly increases from 0.094 to 0.113 because RGPO expands the overall score range. After normalization, however, the relative variation decreases from 0.206 to 0.085, corresponding to a 58.7\% reduction. This shows that RGPO improves robustness to prompt paraphrases while preserving the expanded score resolution.
These results complement the protocol analysis in Appendix A.1. Rubrics with soft-logits reduce score-range collapse before training, while RGPO further improves both score alignment and robustness to prompt paraphrases.

\subsection{EvalVerse 16-Dimensional Taxonomy}
\label{sec:appendix_dim}

Throughout the paper, we adopt the sixteen fine-grained secondary dimensions of the EvalVerse suite, which systematically cover the six coarse axes used in generative video evaluation: \emph{Visual Concept Design}, \emph{Acting}, \emph{Cinematography}, \emph{Aesthetics}, \emph{Affectivity}, and \emph{Multi-Shot Cutting}. Table~\ref{tab:evalverse_dims} provides the definitions and evaluation focus of each dimension. Rubric generation and scoring in RewardVerse are conditioned on the target dimension $d$, enabling the generator to learn dimension-adaptive themes and weights rather than a single generic scoring policy.

\vspace{-0.5cm}

\begin{table}[h]
\centering
\caption{The sixteen secondary evaluation dimensions adapted from the EvalVerse test suite, organized by their high-level axis and specific evaluation focus. All training and testing in Sections~\ref{sec:phenomenon} and~\ref{sec:experiments} are conducted using this taxonomy.}
\label{tab:evalverse_dims}
\small
\setlength{\tabcolsep}{4pt}
\begin{tabular}{lll p{7.5cm}}
\toprule
\textbf{ID} & \textbf{Dimension} & \textbf{Axis} & \textbf{Evaluation Focus} \\
\midrule
\multicolumn{4}{l}{\emph{1. Visual Concept Design}} \\
$D_{1}$  & Character   & Visual Concept & Clarity, recognizability, and adherence of character appearance to the designated settings. \\
$D_{2}$  & Scene       & Visual Concept & Plausibility and appeal of the scene setup; establishment of a clear, distinct, and aesthetically valuable visual style. \\
\midrule
\multicolumn{4}{l}{\emph{2. Acting}} \\
$D_{3}$  & Consistency & Acting         & Coherence of character identities and attributes across frames without temporal drifting, morphing, or flickering. \\
$D_{4}$  & Action      & Acting         & Naturalness and physical plausibility of character movements aligned with their established identity. \\
$D_{5}$  & Expression  & Acting         & Accuracy and vividness of facial expressions in conveying the intended emotions. \\
\midrule
\multicolumn{4}{l}{\emph{3. Cinematography}} \\
$D_{6}$  & Composition & Cinematography & Clarity, stability, and explicit visual intent of individual shot compositions. \\
$D_{7}$  & Lens        & Cinematography & Reasonableness, clarity, and stability of camera parameters (focal length, depth-of-field, etc.). \\
$D_{8}$  & Pacing      & Cinematography & Appropriateness of camera movements and blocking; absence of jarring or unmotivated motion. \\
\midrule
\multicolumn{4}{l}{\emph{4. Aesthetics}} \\
$D_{9}$  & Visual Quality & Aesthetics  & Low-level perceptual quality, temporal consistency, and absence of digital artifacts. \\
$D_{10}$ & Chromaticity   & Aesthetics  & Color harmony, saturation control, and emotive power. \\
$D_{11}$ & Materiality    & Aesthetics  & Aesthetic appeal and self-consistent realism of textures, surface properties, and fine details. \\
$D_{12}$ & Lighting       & Aesthetics  & Physical plausibility and artistic execution of illumination, shading, and cinematic lighting logic. \\
\midrule
\multicolumn{4}{l}{\emph{5. Affectivity}} \\
$D_{13}$ & Grounding   & Affectivity    & Stability and recognizability of the overall emotional direction and atmospheric setting. \\
$D_{14}$ & Progression & Affectivity    & Natural and controlled build-up, transition, and resolution of emotion across the temporal dimension. \\
\midrule
\multicolumn{4}{l}{\emph{6. Multi-Shot Cutting}} \\
$D_{15}$ & Logic       & Multi-Shot     & Continuity of time, space, and action between shot transitions; effectiveness of editing in aiding narrative comprehension. \\
$D_{16}$ & Rhythm      & Multi-Shot     & Overall control of shot durations and cutting cadence; effective emotional guidance and clear temporal organization. \\
\bottomrule
\end{tabular}
\vspace{-0.3cm}
\end{table}

For each dimension, we manually curate a small seed pool of 30 preference pairs $(q, y_w, y_l)$, totaling 480 pairs across the entire taxonomy. These pairs serve both to self-evolve the seed rubric $R^\star_d$ (Section~\ref{sec:warmup}) and to train the RGPO framework. All test-set correlations reported in Table~\ref{tab:pointwise} are evaluated on a disjoint pointwise hold-out set containing 1K evaluation samples ($n = 994$) across the taxonomy.

\paragraph{Baseline Evaluation and Dimension Alignment.}
\label{app:baseline_alignment}

We briefly introduce the external reward models used in Table~\ref{tab:pointwise}. VideoScore-v1.1 is a discriminative video evaluator with five regression heads for visual quality, temporal consistency, motion, text alignment, and factual consistency. VideoScore2 is a generative video evaluator that scores visual quality, text alignment, and physical/common-sense consistency. UnifiedReward is an MLLM-based reward model that evaluates generated content from Style, Physics, and Alignment. VideoReward is a discriminative video
reward model trained with human preferences and predicts visual quality, motion quality, and text alignment. VisionReward is a fine-grained reward model that evaluates videos using a fixed weighted checklist. Q-Scorer is a state-of-the-art IQA model that is particularly strong at structural quality assessment.

We use the official checkpoints without additional training. Since these models do not cover all 16 EvalVerse dimensions, we report the dimensions supported in Table~\ref{tab:pointwise} and mark the others as ``--''. For models with coarse outputs, we use the closest native score following Table~\ref{tab:baseline_dimension_alignment}.

\begin{table*}[t]
\centering
\scriptsize
\caption{Baseline scoring protocols and dimension alignment used in Table~1.}
\label{tab:baseline_dimension_alignment}
\setlength{\tabcolsep}{5pt}
\renewcommand{\arraystretch}{1.12}
\begin{tabular}{
    p{0.14\textwidth}
    p{0.78\textwidth}}
\toprule
Method & Scoring protocol and alignment \\
\midrule

VideoScore-v1.1
&
The model predicts visual quality, temporal consistency, dynamic degree,
text-to-video alignment, and factual consistency.
Visual quality is used for Visual Quality, Chromaticity, Materiality, and
Lighting; temporal consistency for Consistency and Rhythm; dynamic degree
for Action; text-to-video alignment for Scene and Character; and factual
consistency for Logic. \\

VideoScore2
&
The model generates visual quality, text-to-video alignment, and
physical/common-sense consistency scores.
Visual quality is used for Visual Quality, Chromaticity, Materiality, and
Lighting; text-to-video alignment for Scene and Character; and
physical/common-sense consistency for Consistency, Action, Logic, and
Rhythm. \\

UnifiedReward
&
We use its Style, Physics, and Alignment scores.
Style is used for Visual Quality, Chromaticity, Materiality, Lighting, and
Character; Physics for Consistency, Action, and Rhythm; and Alignment for
Scene and Logic. \\

VideoReward
&
We use its visual quality (VQ), motion quality (MQ), and text alignment
(TA) scores.
VQ is used for Visual Quality, Chromaticity, Materiality, Lighting, and
Character; MQ for Consistency, Action, and Rhythm; and TA for Scene and
Logic. \\

VisionReward
&
We use the score from its fixed video checklist for all reported
dimensions. The checklist is not changed across dimensions. \\

Q-Scorer
&
We sample video frames, score each frame using the official image-quality
prompt, and average the frame scores. The same frame-averaged score is used
for all reported dimensions. \\

\bottomrule
\end{tabular}
\vspace{-0.5cm}
\end{table*}

\subsection{DOWNSTREAM GRPO FINE-TUNING}
\label{app:downstream}

\subsubsection{FINE-TUNING SETUP AND EVALUATION}

We further evaluate whether RewardVerse can serve as an effective reward signal for downstream video generation optimization. We fine-tune Wan-2.2-A14B with GRPO using either VideoReward or RewardVerse as the reward model, while the original Wan-2.2-A14B is used as the base model. The optimization targets only the \textit{Visual Quality} dimension. All training settings are kept identical across the two GRPO runs. We train with LoRA rank 64 and a rollout group size of $G=16$.
For evaluation, we use the VBench prompt suite \citep{huang2024vbench} and generate 17-frame videos at $480\times832$ resolution with deterministic sampling. We report VBench-Quality, averaged over six official quality dimensions, and VBench-Text, averaged over three official text-related dimensions, together with several representative VBench dimensions. Since only Visual Quality is optimized, the remaining dimensions are used to examine whether reward optimization degrades other video capabilities.

\subsubsection{DOWNSTREAM RESULTS}

\begin{table}[t]
\centering
\caption{
Downstream GRPO fine-tuning of Wan-2.2-A14B on the Visual Quality dimension.
RewardVerse improves the target quality metric while largely preserving other video capabilities.
Best results are shown in bold.
}
\label{tab:downstream}
\begin{tabular}{lccc}
\toprule
Metric & Base & VideoReward & RewardVerse \\
\midrule
Imaging Quality        & 0.640 & 0.648 & \textbf{0.653} \\
Human Action           & 0.900 & 0.800 & \textbf{0.950} \\
Background Consistency & 0.919 & 0.916 & \textbf{0.920} \\
Temporal Flickering    & 0.943 & 0.938 & \textbf{0.944} \\
Motion Smoothness      & \textbf{0.966} & 0.960 & \textbf{0.966} \\
Subject Consistency    & \textbf{0.898} & 0.881 & 0.894 \\
\midrule
VBench-Quality (6D)           & 0.808 & 0.804 & \textbf{0.809} \\
VBench-Text (3D)         & 0.428 & 0.392 & \textbf{0.446} \\
\bottomrule
\end{tabular}
\vspace{-0.5cm}
\end{table}

As shown in Table~\ref{tab:downstream}, RewardVerse improves Imaging Quality from 0.640 to 0.653, which is the dimension most directly related to the optimized Visual Quality reward. It also improves Human Action from 0.900 to 0.950, while Background Consistency, Temporal Flickering, Motion Smoothness, and Subject Consistency remain close to the base model. In contrast, optimization with VideoReward reduces both VBench-Quality from 0.808 to 0.804 and VBench-Text from 0.428 to 0.392.

These results show that RewardVerse improves the target visual quality without clear degradation of other video capabilities. In comparison, VideoReward improves Imaging Quality but reduces the overall VBench performance, suggesting that RewardVerse is less prone to reward hacking during downstream optimization.

\subsection{Prompts and Training Details}
\label{app:training_details}

\subsubsection{Prompt Templates}
\label{app:prompts}

\paragraph{Seed Rubric Generation.}
For each training pair, the teacher receives the target dimension, generation prompt, videos, and human scores. It is asked to generate observable and dimension-specific criteria, with all rubric weights summing to one. The generated rubric is then used to score the videos and is revised when its prediction differs from the human label by more than $0.5$. We use at most
five revision rounds and remove near-duplicate rubrics using a token-level Jaccard threshold of $0.85$.

\paragraph{Dynamic Rubric Generation.}
The rubric generator uses the following prompt:

\begin{promptbox}
You are a cinematic video evaluation expert. Generate exactly five rubric items for scoring videos on the {dimension} dimension.

Task:
{task_description}

Caption:
{caption}

Each rubric contains:
- a clear theme;
- a weight in (0, 1), with all weights summing to 1;
- at least two observable evaluation tips.

Output strict JSON:
{
  "rubrics": [
    {
      "theme": "...",
      "weight": 0.NN,
      "tips": ["...", "..."]
    },
    ...
  ]
}
\end{promptbox}

\paragraph{Rubric-Guided Scoring.}
Given a video and a generated rubric, the scorer uses the following prompt:

\begin{promptbox}
You are a strict video reward judge for the {dimension} dimension.

Score the single video on each rubric using an integer from 1 to 5.

Caption:
{caption}

Rubric:
{rubric_text}

Respond with exactly one line:
theme_1: _ theme_2: _ theme_3: _ theme_4: _ theme_5: _
\end{promptbox}

At each score position, we read the logits of tokens ``1'' to ``5'' and compute their expected value. This fixed output format provides continuous theme scores without parsing free-form text.

\subsubsection{Reward Model Training Settings}
\label{loss}

We train RewardVerse on Qwen2.5-VL-7B using the 480 preference pairs described in Appendix~\ref{sec:appendix_dim}. The training is divided into two separate runs corresponding to the two stages of RGPO. Both the scorer warm-up and joint optimization are trained for one epoch. Each epoch contains approximately 60 optimization steps. For each training pair, we sample a rollout group of size $G=8$.

For both stages, we use AdamW with a learning rate of $2\times10^{-6}$, 30 learning-rate warm-up steps, a preference temperature of $1.5$, and a maximum gradient norm of $1.0$. The GRPO KL coefficient is set to $0.1$. Training uses bfloat16 precision and DeepSpeed ZeRO-3 on 8 NVIDIA H20 GPUs.

\paragraph{Stage 1: Scorer Warm-up.}
The student and reference policies are initialized from Qwen2.5-VL-7B. The seed rubric is fixed for all rollouts, so this stage only updates the scorer role. The scorer update coefficient is fixed at $0.4$ throughout the epoch. We set the margin calibration weight
$\lambda_{\mathrm{cal}}$ to $0.2$, the score-format gate to $0.55$, and use an additional score-side KL coefficient of $0.005$ to limit score-scale drift. The model is trained for one epoch.

\paragraph{Stage 2: Joint Optimization.}
We initialize the student from the final Stage-1 checkpoint. The same checkpoint is used as the frozen reference policy for both rubric generation and scoring. The generator produces dynamic rubrics from the query, and the scorer evaluates both videos under each sampled rubric. The rubric-alignment weight and format reward weight are both fixed at $0.5$. We increase the margin calibration weight to $0.3$ and use a score-side KL coefficient of $0.015$. The scorer update coefficient is linearly increased from zero at step 20 to $0.4$ at step 30. Joint optimization is also run for one epoch. The Stage-2 checkpoint at step 45 is selected based on the macro-averaged PLCC on the held-out validation set and used for all evaluations.

\paragraph{Stage-2 GRPO Objective.}
For completeness, the generator-side GRPO objective used in Stage 2 is:
\begin{equation}
\begin{split}
    \mathcal{L}_{\mathrm{GRPO}}^{(2)}(\theta)
    = -\frac{1}{G}\sum_{g=1}^{G}\frac{1}{|\hat{R}_g|}
    \sum_{t=1}^{|\hat{R}_g|}
    \Bigl[
    &\min\!\bigl(
    \rho_{g,t}(\theta)\hat{A}_g,\,
    \mathrm{clip}(\rho_{g,t}(\theta),1{-}\epsilon,1{+}\epsilon)\hat{A}_g
    \bigr) \\
    &-\beta_{\mathrm{KL}}\,
    \mathbb{D}_{\mathrm{KL}}\!\bigl[
    \pi_\theta^{\mathrm{gen}}
    \,\big\|\,
    \pi_{\mathrm{ref}}^{\mathrm{gen}}
    \bigr]_{g,t}
    \Bigr],
\end{split}
\label{eq:stage2_grpo_full}
\end{equation}
where $\hat{A}_g$ and $\rho_{g,t}(\theta)$ are defined in Section~\ref{sec:joint}, and $\pi_{\mathrm{ref}}^{\mathrm{gen}}$ is the frozen reference generator.

\begin{table}[t]
\centering
\small
\caption{Main hyperparameters for the two-stage RewardVerse training.
Both stages are trained for one epoch.}
\label{tab:rm_training_settings}
\begin{tabular}{lcc}
\toprule
Setting & Stage 1: Warm-up & Stage 2: Joint \\
\midrule
Initialization
    & Qwen2.5-VL-7B
    & Stage-1 checkpoint \\
Reference policy
    & Qwen2.5-VL-7B
    & Stage-1 checkpoint \\
Training epochs
    & 1
    & 1 \\
Optimization steps
    & $\sim$60
    & $\sim$60 \\
Rollouts per sample $G$
    & 8
    & 8 \\
Learning rate
    & $2\times10^{-6}$
    & $2\times10^{-6}$ \\
LR warm-up steps
    & 30
    & 30 \\
Preference temperature
    & 1.5
    & 1.5 \\
GRPO KL coefficient
    & 0.1
    & 0.1 \\
Score-side KL coefficient
    & 0.005
    & 0.015 \\
Margin calibration weight
    & 0.2
    & 0.3 \\
Rubric-alignment weight
    & -- 
    & 0.5 \\
Format reward weight 
    & -- 
    & 0.5 \\
Maximum scorer update weight
    & 0.4
    & 0.4 \\
Scorer update schedule
    & Fixed
    & Steps 20--30 \\
Score-format gate
    & 0.55
    & 0.55 \\
Maximum gradient norm
    & 1.0
    & 1.0 \\
\bottomrule
\end{tabular}
\end{table}

\subsubsection{Qualitative Study: Inside the Learned Rubrics}
\label{app:qualitative}

\begin{figure}[t]
\centering
\includegraphics[width=1.0\linewidth]{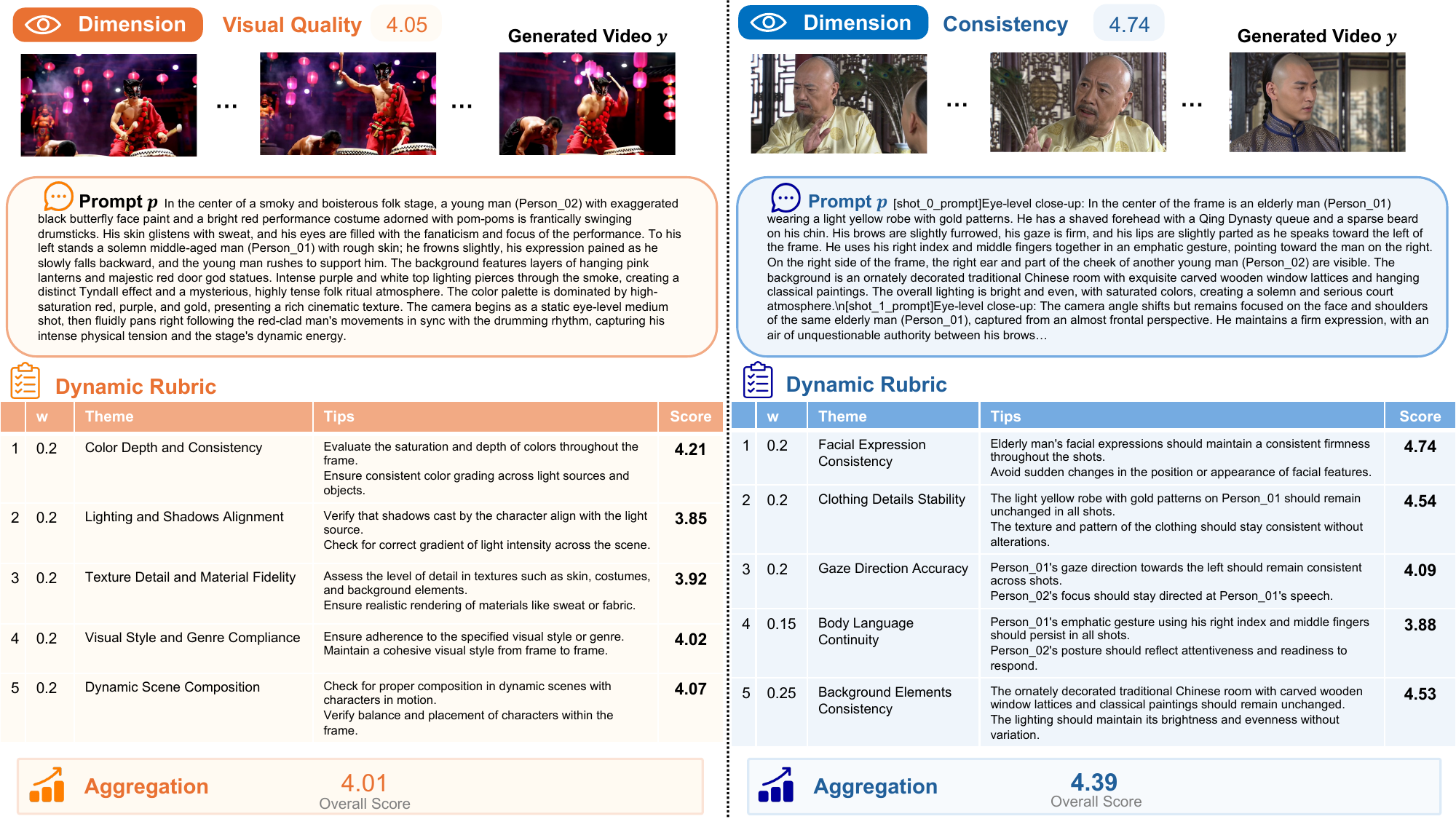}
\vspace{-0.3cm}
\caption{Qualitative view of dynamic rubrics generated by RewardVerse. Left (\textit{Visual Quality}): themes and tips stay abstract. Right (\textit{Consistency}): tips dynamically ground on specific entities in the prompt, illustrating scene-adaptation.}
\label{fig3}
\vspace{-0.5cm}
\end{figure}

Figure~\ref{fig3} visualizes the learned rubrics generated by RewardVerse across different prompts and evaluation dimensions. Notably, the generator automatically adjusts the level of abstraction in the tips based on the target dimension. For global, technical, or aesthetic axes such as \textit{Visual Quality}, the generated tips remain abstract to avoid overfitting to the specific prompt content. In contrast, for identity-bound or scene-bound axes such as \textit{Consistency}, the tips dynamically ground on specific entities in the prompt (e.g., monitoring ``the light yellow robe with gold patterns on Person\_01''), as these entities are exactly what must remain stable. This example illustrates how the generator adjusts the tip granularity to different evaluation dimensions. 

\subsection{Inference-Time Cost}
\label{app:cost}

We measure the inference latency of the four pointwise scoring variants on a single H20 under the same inference settings (bf16, greedy decoding, FPS = 2, and a 64-frame budget). All variants use the same visual backbone and frame-sampling pipeline.

As shown in Table~\ref{tab:cost}, rubric generation introduces only a
small inference overhead. V3 takes approximately 2.0 s per video, compared with 1.7 s for the rubric-free V2 baseline, corresponding to an increase of about 18\%. V4 is slightly slower because it additionally generates multiple theme scores in natural text.

The rubric depends only on the evaluation query rather than the video. Therefore, when multiple candidate videos are evaluated under the same query, the rubric can be generated once and reused, further reducing the effective per-video overhead.

\begin{table}[h]
\centering
\caption{Per-video inference latency of the four pointwise scoring variants on a single H20.}
\label{tab:cost}
\begin{tabular}{lccc}
\toprule
Variant & Rubric & Readout & Latency \\
\midrule
V1 (no rubric, NL float)              & \xmark     & NL float    & $\approx 1.7$ s \\
V2 (no rubric, soft-logits)           & \xmark     & soft-logits & $\approx 1.7$ s \\
V3 (rubric + soft-logits, deployed)   & \checkmark & soft-logits & $\approx 2.0$ s \\
V4 (rubric + multi-theme NL float)    & \checkmark & NL float    & $\approx 2.2$ s \\
\bottomrule
\end{tabular}
\vspace{-0.5cm}
\end{table}

\subsection{Limitations and Future Work}
\label{app:limitations}

Our current results should be read with a few caveats in mind.

\paragraph{Backbone Capacity.} For a controlled and fair evaluation against existing video reward models, we intentionally selected the standard-sized Qwen2.5-VL-7B as our backbone. However, the effectiveness of RGPO is bounded by the capability of the underlying MLLM. RGPO can better align the model’s existing evaluation ability with human judgments, but it cannot recover visual or temporal details that the backbone fails to recognize. Therefore, stronger backbones are expected to produce more reliable rubrics and scores, leading to a higher performance ceiling for RL optimization.

\paragraph{Theme-Level Calibration.} Our current calibration objective directly aligns only the aggregated score margin between preferred and non-preferred videos. The score of each individual rubric theme is not directly supervised. As a result, the final reward can be well aligned with human preferences while some theme scores remain biased or inconsistent. Directly calibrating theme-level scores is therefore an important direction for improving both score reliability and interpretability.

\paragraph{Scaling.} Our current experiments are conducted with a 7B reward backbone, only 30 preference pairs per dimension, and one 14B video generation model for downstream RL. It remains necessary to scale RewardVerse to stronger reward backbones, larger training sets, and more video generation models to verify whether the observed gains remain consistent at scale.

\paragraph{Relation to CoT-Based Scoring.}
Recent CoT-based visual scoring methods treat reasoning as an intermediate representation that summarizes visual evidence before producing a quality score~\citep{zhao2025reasoning, he2025videoscore2, wang2026think}. RewardVerse operates at a different stage. Its rubric is generated from the evaluation dimension and prompt without video access, and specifies what should be checked before scoring. Therefore, CoT summarizes the evidence observed in the video, whereas the rubric defines the evaluation criteria. The two designs are complementary, and the generated rubric may further guide a CoT-based scorer.

\paragraph{Future Work.} Future work will focus on stronger MLLM backbones, direct theme-level calibration, larger-scale training and downstream RL experiments, and rubric-guided CoT scoring.

\end{document}

%% file: math_commands.tex
\usepackage{amsmath,amsfonts,bm}

\def\eqref#1{equation~\ref{#1}}
\def\1{\bm{1}}

\DeclareMathAlphabet{\mathsfit}{\encodingdefault}{\sfdefault}{m}{sl}
\SetMathAlphabet{\mathsfit}{bold}{\encodingdefault}{\sfdefault}{bx}{n}